\documentclass{article} % For LaTeX2e
\usepackage{iclr2026_conference,times}

\usepackage{amsmath,amsfonts,bm}

\def\eqref#1{equation~\ref{#1}}
\def\1{\bm{1}}

\DeclareMathAlphabet{\mathsfit}{\encodingdefault}{\sfdefault}{m}{sl}
\SetMathAlphabet{\mathsfit}{bold}{\encodingdefault}{\sfdefault}{bx}{n}

\usepackage{hyperref}
\usepackage{url}

\usepackage[utf8]{inputenc} % allow utf-8 input
\usepackage[T1]{fontenc}    % use 8-bit T1 fonts
\usepackage{hyperref}       % hyperlinks
\usepackage{url}            % simple URL typesetting
\usepackage{booktabs}       % professional-quality tables
\usepackage{amsfonts}       % blackboard math symbols
\usepackage{nicefrac}       % compact symbols for 1/2, etc.
\usepackage{microtype}      % microtypography
\usepackage{xcolor}         % colors

\usepackage{graphicx}
\usepackage{multirow}
\usepackage[table]{xcolor}
\usepackage{tabularx} % 用于设置表格总宽度
\usepackage{makecell} % 用于单元格内换行
\usepackage{amsmath}
\usepackage{bbm} % for \mathbbm{1}
\usepackage{booktabs}
\usepackage{multirow}
\usepackage{subcaption}
\usepackage{wrapfig}
\usepackage{amssymb}
\usepackage{mathtools}

\title{\textsc{PixWorld}: Unifying 3D Scene Generation and Reconstruction in Pixel Space}

\author{
Sensen Gao$^{1*}$, Zhaoqing Wang$^{2*}$, Qihang Cao$^{1}$, Dongdong Yu$^{2}$, Changhu Wang$^{2}$, \\
\textbf{Jia-Wang Bian}$^{2\dagger}$  \\
  $^{1}$ Nanyang Technological University \quad 
  $^{2}$ AISphere  \\
  * Co-first authors. \quad $^{\dagger}$Corresponding authors.\\
}

\iclrfinalcopy % Uncomment for camera-ready version, but NOT for submission.
\begin{document}

\maketitle

\vspace{-8mm}
\begin{figure*}[h]
    \centering
    \includegraphics[width=0.95\textwidth]{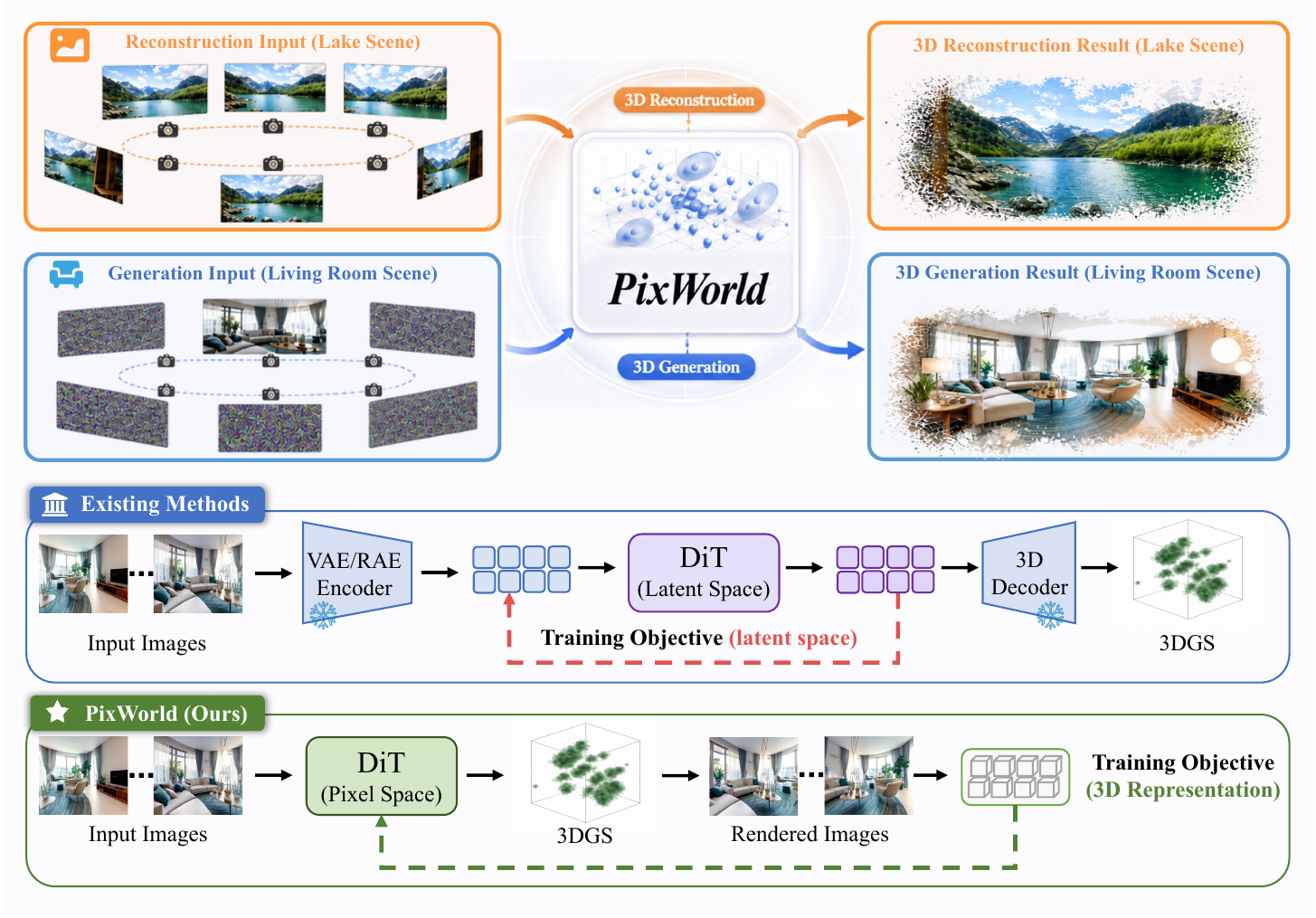}
    \vspace{-4mm}
    \caption{PixWorld unifies 3D scene reconstruction and generation within a single model. Unlike prior approaches that compute losses in the latent space of a VAE~\citep{kingma2013auto} or RAE~\citep{zheng2025diffusion}, PixWorld applies a flow matching objective directly in pixel space over multi-view renderings, enabling end-to-end optimization of the underlying 3D representation. This design avoids the information loss inherent to latent representations and eliminates the cost of pretraining a VAE or RAE.}
    \label{fig:teaser}
\end{figure*}

\begin{abstract}
3D reconstruction and generation are commonly tackled by separate paradigms: pixel-based regression for reconstruction, and latent diffusion for generation. Recent works attempt to unify them in latent space, but with notable drawbacks: the diffusion objective is defined on latent features rather than the underlying 3D representation, and both branches suffer from information loss introduced by latent encoding, while requiring a pretrained Variational Autoencoder (VAE) or Representation Autoencoder (RAE). In this paper, we reformulate these two tasks under a unified pixel-space diffusion paradigm and introduce PixWorld, a single model that jointly addresses 3D reconstruction and generation. By supervising diffusion directly on rendered images, PixWorld removes the above limitations and aligns optimization with 3D scene fidelity. Beyond photometric and perceptual supervision that operates at the 2D image level and lacks 3D geometric awareness, we further introduce a geometry perception loss that aligns rendered views with their ground truth in the geometry-aware feature space of a pretrained 3D foundation model, providing 3D structural supervision. PixWorld consistently outperforms prior latent-space generation methods and matches state-of-the-art reconstruction methods, demonstrating the superiority of a unified pixel-space approach.
\end{abstract}

\section{Introduction}

Building 3D scenes from visual observations is a long-standing goal in computer vision, with broad impact across gaming, robotics, embodied AI, and VR/AR~\citep{ding2025understanding,kong20253d,ye2026world,zhang2025advances}. Two complementary directions have driven progress in this space: reconstruction recovers 3D scenes from real-world captures, while generation synthesizes plausible scenes from limited or even imagined conditions~\citep{zhang2025advances,wang2026feed,li2026thinking}. Together, they form the technological foundation for populating the digital worlds of tomorrow.

3D scene generation and reconstruction have long developed as two separate lines of research. Reconstruction is dominated by feed-forward methods that regress 3D representations directly from multi-view images~\citep{hong2023lrm,charatan2024pixelsplat,szymanowicz2024splatter,xu2025depthsplat,chen2024mvsplat}. Generation has evolved from per-scene optimization with 2D priors via score distillation~\citep{lin2023magic3d,poole2022dreamfusion,tang2023dreamgaussian,wang2023prolificdreamer} to latent-space diffusion as the current mainstream~\citep{yang2025prometheus,go2025splatflow,huang2026gen3r,li2025flashworld}, with recent extensions diffusing in the feature space of pretrained 3D foundation models or representation autoencoders~\citep{gao2026oneworld,sun2026vggt,jang2026repurposing}. A recent work, Gen3R~\citep{huang2026gen3r}, attempts to unify the two tasks by extending the latent-space generation pipeline to also handle reconstruction. However, this design introduces clear limitations: the diffusion objective is defined on intermediate latent features rather than the underlying 3D representation, preventing the 3D output from being directly optimized; in addition, both branches suffer from information loss introduced by the pretrained Variational Autoencoder (VAE) or Representation Autoencoder (RAE), which itself requires additional training.

To address these limitations, we reformulate unification under a pixel-space diffusion paradigm and introduce PixWorld, a single framework that jointly handles 3D reconstruction and generation. This design brings two key benefits. First, for both generation and reconstruction, eliminating the latent stage removes the information loss introduced by latent encoding and the additional training cost of a VAE/RAE, which is especially critical for reconstruction as a fidelity-bound task. Second, for generation, the diffusion objective directly supervises the 3D representation through differentiable rendering (see Fig.~\ref{fig:teaser}), aligning the training signal with the fidelity of the 3D scene rather than with targets in an intermediate latent space. Building on this pixel-space framework, the unification reduces to how to expose both tasks to a single forward pass. We achieve this by partitioning multi-view inputs into clean and noisy subsets: clean views drive reconstruction, while noisy ones are generated conditioned on the clean ones, with both producing a pixel-aligned 3D Gaussian representation~\citep{kerbl20233d}.

Despite this direct pixel-space supervision, image-level photometric and perceptual losses do not fully guarantee geometrically faithful 3D structure.
To address this, we introduce a geometry perception loss that brings rendered views and their ground-truth counterparts close in the geometry-aware feature space of a pretrained 3D foundation model. As this feature space encodes 3D geometric structure beyond 2D appearance, the loss directly encourages the rendered scene to share the same underlying geometry as the ground truth. Extensive experiments on RealEstate10K~\citep{zhou2018stereo}, DL3DV~\citep{ling2024dl3dv} and WorldScore~\citep{duan2025worldscore} demonstrate that a single PixWorld model delivers high-fidelity 3D scene generation and reconstruction, matching state-of-the-art reconstruction methods while outperforming previous latent-space generation approaches.

Our main contributions are summarized as follows:
\begin{itemize}
\item We propose PixWorld, an end-to-end pixel-space diffusion framework that supervises a pixel-aligned 3D Gaussian representation directly through multi-view differentiable rendering, with no intermediate VAE or RAE. This eliminates the information loss and training cost of a latent autoencoder, aligns the diffusion signal with the fidelity of the 3D scene itself, and naturally unifies 3D scene generation and reconstruction in a single model.
\item We introduce a geometry perception loss that aligns rendered views with ground truth in the geometry-aware feature space of a pretrained 3D foundation model, providing 3D structural supervision beyond 2D photometric and perceptual losses.
\item Experiments show that a single PixWorld model achieves superior performance on both 3D scene generation and reconstruction, establishing pixel-space diffusion as a unified paradigm for 3D scene modeling.
\end{itemize}

\section{Related Work}

\subsection{3D Scene Generation}
\textbf{Diffusion-based Iterative 3D Scene Generation.} 
Early 3D generation methods leverage pretrained 2D generative models~\citep{rombach2022high,podell2023sdxl,zhang2023adding,peebles2023scalable} as generative priors. A representative line employs Score Distillation Sampling (SDS)~\citep{lin2023magic3d,poole2022dreamfusion,tang2023dreamgaussian,wang2023prolificdreamer} to optimize a 3D representation such as 3DGS~\citep{kerbl20233d} or NeRF~\citep{mildenhall2021nerf} by aligning rendered views with a pretrained 2D diffusion model. Without explicit multi-view constraints, these methods suffer from cross-view inconsistency, and the per-scene iterative optimization further limits their scalability.

\textbf{Multi-View Reconstruction-Based 3D Scene Generation.}
A second line first synthesizes multi-view images or videos with pretrained 2D diffusion models, then performs 3D reconstruction from the synthesized views~\citep{chen2024mvsplat360,gao2024cat3d,hao2025gaussvideodreamer,liu2024reconx,liu2023syncdreamer,sargent2023zeronvs,shi2023mvdream,sun2024dimensionx,wu2024reconfusion,zhao2024genxd} or incremental outpainting~\citep{chung2023luciddreamer,fridman2023scenescape,schwarz2025recipe,yu2024wonderjourney,yu2025wonderworld}. While avoiding per-scene optimization, these methods rely on 2D RGB priors without explicit 3D reasoning, often resulting in inconsistent geometry and weak multi-view fidelity.

\textbf{3D Scene Generation in Latent Spaces.}
More recent work moves diffusion into compressed latent spaces and learns a latent-to-3D decoder, typically targeting 3DGS. One line freezes a pretrained image or video VAE (\textit{e.g.}, SD-VAE~\citep{podell2023sdxl}, Wan-VAE~\citep{wan2025wan}) and trains a decoder that maps the resulting latents to 3D Gaussians, with a diffusion backbone fine-tuned in this latent space~\citep{yang2025prometheus,go2025splatflow,li2024director3d,li2025flashworld,dai2025fantasyworld,go2025videorfsplat,go2025vist3a,zhou2026drivinggen,wang2025mmgen}. Following the shift from VAEs to Representation Autoencoders (RAEs)~\citep{zheng2025diffusion,shi2025latent,chen2025aligning,bi2025vision}, a more recent line operates in the representation space of pretrained 3D foundation models~\citep{gao2026oneworld,sun2026vggt,jang2026repurposing}. Building on this latent-space paradigm, Gen3R~\citep{huang2026gen3r} further attempts to unify reconstruction and generation within a single latent-space model.

\subsection{3D Scene Reconstruction}
3D scene reconstruction aims to recover 3D representations from one or more captured views. Classical pipelines rely on multi-view stereo and structure-from-motion to estimate dense geometry, while NeRF~\citep{mildenhall2021nerf} and 3D Gaussian Splatting~\citep{kerbl20233d} popularized per-scene optimization for high-fidelity novel view synthesis. To avoid expensive per-scene fitting, recent feed-forward methods learn to directly map sparse or dense multi-view inputs to a 3D representation in a single forward pass~\citep{hong2023lrm,charatan2024pixelsplat,szymanowicz2024splatter,chen2024mvsplat,xu2025depthsplat}, with LRM~\citep{hong2023lrm} regressing implicit triplanes and pixel-aligned Gaussian regressors~\citep{charatan2024pixelsplat,szymanowicz2024splatter,chen2024mvsplat} predicting per-pixel Gaussians. These methods are deterministic and conditioned on the available observations, and therefore cannot synthesize unseen content beyond the input views.

\subsection{Pixel-Space Generation}
Latent diffusion models~\citep{rombach2022high,podell2023sdxl,peebles2023scalable} dominate large-scale image generation by operating in a compressed VAE latent space, introducing an indirection between the generative target and the pixel output. A complementary line of work performs diffusion directly in pixel space~\citep{dhariwal2021diffusion,jabri2023scalable,hoogeboom2023simple,hoogeboom2025simpler,wang2025pixnerd,chen2025pixelflow,yu2025pixeldit,ma2025deco,chen2025dip,li2025back}, showing that, with sufficient capacity and data, pixel-space generation can match or surpass its latent-space counterparts in fidelity. We extend this perspective to the 3D setting: by performing diffusion directly in pixel space, PixWorld removes the intermediate VAE/RAE of prior latent-space 3D methods and supervises the 3D representation directly through differentiable rendering of the predicted 3DGS.

%======================================================================
\section{Methodology}
\label{sec:method}
%======================================================================

\begin{figure}[t]
    \centering
    \includegraphics[width=\textwidth]{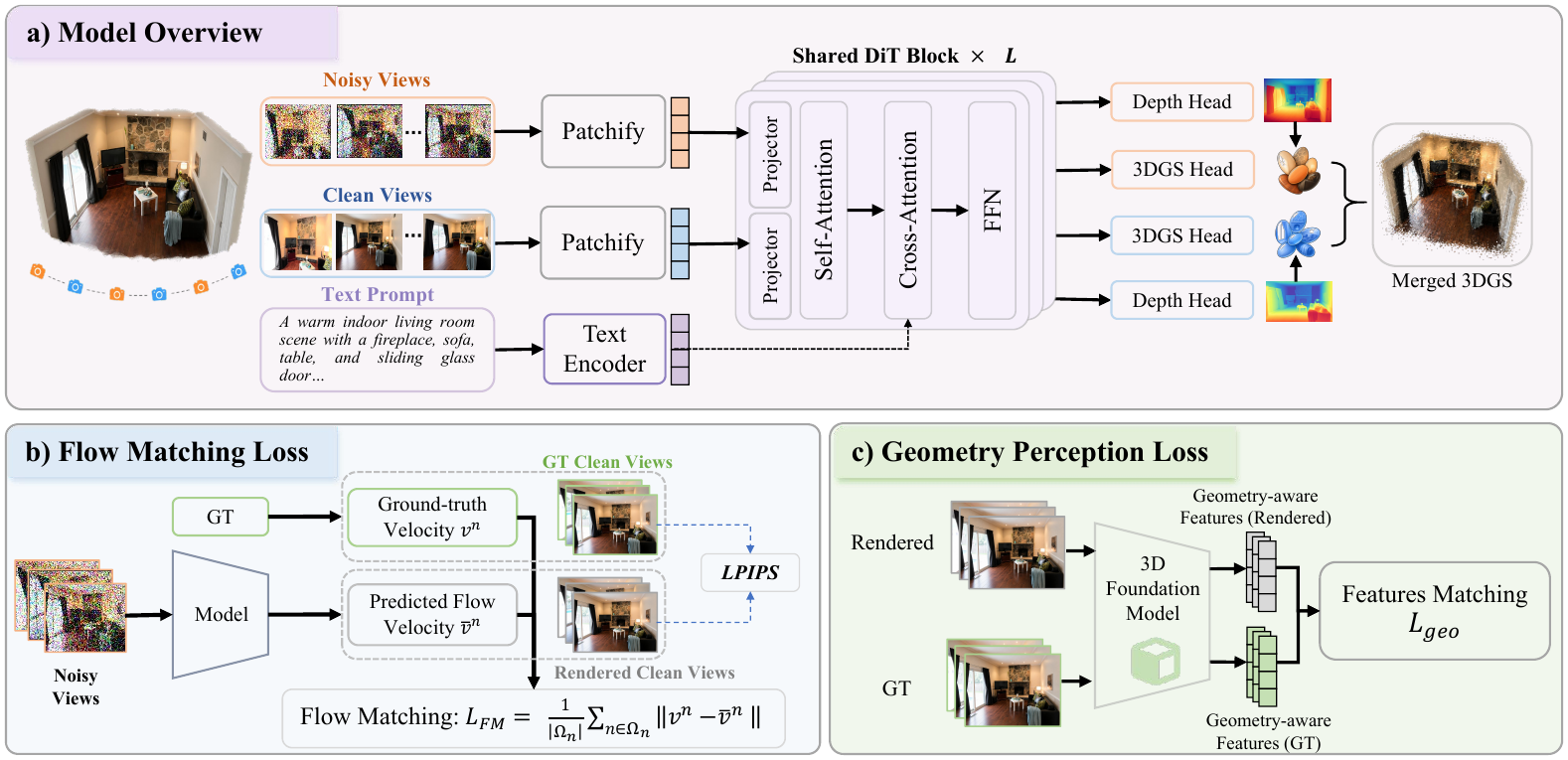}
    \caption{\textbf{Overview of PixWorld.} (a) PixWorld adopts a unified DiT-based framework that takes noisy and clean multi-view inputs, with optional text conditioning, and jointly predicts depth and 3DGS through shared transformer blocks. (b) A pixel-space flow matching loss is imposed on rendered multi-view images to directly optimize the underlying 3D representation. (c) A geometry perception loss further enforces structural consistency by aligning rendered views with ground-truth observations through a 3D foundation model.}
    \vspace{-6mm}
    \label{fig:method}
\end{figure}

We present PixWorld, a unified pixel-space diffusion framework that jointly addresses 3D scene reconstruction and generation within a single model (Fig.~\ref{fig:method}). Given a set of posed multi-view images, PixWorld partitions them into a clean subset and a noisy subset, jointly processes both with a two-stream diffusion transformer, and decodes the resulting features into a pixel-aligned 3D Gaussian representation~\citep{kerbl20233d}. The predicted Gaussians are rendered back to images, on which the diffusion objective is supervised directly, aligning optimization with 3D scene fidelity rather than with targets in an intermediate latent space. To further provide 3D structural supervision beyond photometric and perceptual losses that operate at the 2D image level, we introduce a geometry perception loss defined in the geometry-aware feature space of a pretrained 3D foundation model.

%======================================================================
\subsection{Preliminary: Pixel-Space Diffusion}
\label{sec:prelim}
%======================================================================

We first briefly review diffusion in the standard 2D image setting and contrast pixel-space diffusion with latent-space diffusion. In latent diffusion, an image $x$ is first mapped into a compact latent representation by a pretrained image autoencoder, \textit{e.g.}, a VAE or a RAE, written as $\mathcal{E}_{\mathrm{AE}}: x \mapsto z$. Diffusion is then performed in the latent space, and the denoised latent is finally mapped back to RGB space by $\mathcal{D}_{\mathrm{AE}}: \hat{z} \mapsto \hat{x}$. This design reduces the dimensionality and computational cost of diffusion modeling, but it also inserts an intermediate autoencoding bottleneck between the diffusion variable and the final image-space supervision.

Following JiT~\citep{li2025back}, we instead adopt \emph{image prediction} in pixel space. Given a clean image $x\in\mathbb{R}^{H\times W\times 3}$, a Gaussian noise sample $\epsilon\sim\mathcal{N}(\mathbf{0},\mathbf{I})$, and a timestep $t\in[0,1]$, the noisy input is constructed by linear interpolation, $x_t = t\,x + (1-t)\,\epsilon$. The denoiser is parameterized directly as an image predictor,
\begin{equation}
f_\theta : (x_t,t,c) \mapsto \hat{x},
\label{eq:prelim_xpred}
\end{equation}
where $c$ denotes conditional information such as class labels or text embeddings. The predicted image is converted into a velocity field $\hat{v}=(\hat{x}-x_t)/(1-t)$, yielding the flow-matching objective
\begin{equation}
\mathcal{L}_{\mathrm{FM}}
=
\mathbb{E}_{x,\epsilon,t}\!\left[\|\hat{v}-v\|_2^2\right]
=
\mathbb{E}_{x,\epsilon,t}\!\left[\frac{\|\hat{x}-x\|_2^2}{(1-t)^2}\right],
\label{eq:prelim_fm}
\end{equation}
where $v=x-\epsilon$ is the ground-truth velocity. This pixel-space parameterization is particularly attractive for our 3D setting. Latent diffusion defines its objective on encoded latents, leaving the diffusion process decoupled from both the rendered output and the underlying 3D representation it is meant to supervise. Pixel-space diffusion instead keeps the diffusion variable and the rendered output in the same RGB domain, so the diffusion objective can be supervised directly on rendered images, aligning optimization with 3D scene fidelity rather than with targets in an intermediate latent space.

%======================================================================
\subsection{PixWorld Framework}
\label{sec:pixworld}
%======================================================================

\paragraph{Task formulation.}
Building on the 2D pixel-space diffusion formulation above, PixWorld extends image prediction to a posed multi-view setting. Given a scene with $N$ posed views, we denote the RGB images and camera parameters by
$
\mathcal{I}=\{\mathcal{I}^n\}_{n=1}^{N},
\mathcal{T}=\{\mathcal{T}^n\}_{n=1}^{N},
$
where $\mathcal{I}^n\in\mathbb{R}^{H\times W\times 3}$ is the RGB image of the $n$-th view and $\mathcal{T}^n$ denotes its camera parameters. We partition the view indices into a clean subset $\Omega_{\mathrm c}$ and a noisy subset $\Omega_{\mathrm n}$ such that
\begin{equation}
\Omega_{\mathrm c}\cup\Omega_{\mathrm n}=\{1,\dots,N\},
\qquad
\Omega_{\mathrm c}\cap\Omega_{\mathrm n}=\varnothing,
\qquad
|\Omega_{\mathrm c}|\ge 1.
\label{eq:clean_noisy_partition}
\end{equation}
When $\Omega_{\mathrm n}=\varnothing$, the task reduces to multi-view 3D reconstruction from clean observations. Otherwise, PixWorld predicts the noisy views conditioned on the clean ones. An optional text prompt $y$ may further specify the scene.

For each view $n$, the model input is defined as
\begin{equation}
\tilde{\mathcal{I}}_t^n=
\begin{cases}
\mathcal{I}^n, & n\in\Omega_{\mathrm c},\\[2pt]
t\,\mathcal{I}^n+(1-t)\epsilon^n, & n\in\Omega_{\mathrm n},
\end{cases}
\qquad
\epsilon^n\sim\mathcal{N}(\mathbf 0,\mathbf I),
\label{eq:mixed_input}
\end{equation}
where all noisy views of the same scene share the same timestep $t$. We denote the mixed multi-view input by $\tilde{\mathcal{I}}_t=\{\tilde{\mathcal{I}}_t^n\}_{n=1}^{N}$.

\paragraph{Two-stream diffusion transformer.}
Given the mixed input $\tilde{\mathcal{I}}_t$, PixWorld uses a two-stream diffusion transformer
\begin{equation}
f_\theta : (\tilde{\mathcal{I}}_t,\mathcal{T},y)\mapsto (\hat{\mathcal{D}},\hat{\mathcal{G}}),
\label{eq:pixworld_mapping}
\end{equation}
where $\hat{\mathcal{D}}=\{\hat{\mathcal{D}}^n\}_{n=1}^{N}$ denotes the predicted multi-view depth maps and $\hat{\mathcal{G}}$ denotes the predicted pixel-aligned 3D Gaussian scene representation. Clean and noisy views are embedded by separate input projections and jointly processed by shared transformer blocks. The noisy stream is conditioned on the sampled timestep $t$, while the clean stream always receives the fully denoised time embedding corresponding to $t=1$. Camera parameters are injected via PRoPE~\citep{li2025cameras}, and the optional text condition is fused through cross-attention.

\paragraph{3D Gaussian decoding and rendering.}
From the shared features, the network predicts a depth map $\hat{\mathcal{D}}^n$ for each view together with the Gaussian attributes that define $\hat{\mathcal{G}}$. Rather than regressing 3D centers directly, we unproject each pixel using its predicted depth and camera: for the $p$-th pixel of the $n$-th view with depth $\hat d_p^n$, the Gaussian center is $\mu_p^n=\Pi^{-1}(p,\hat d_p^n,\mathcal{T}^n)$, where $\Pi^{-1}$ is the inverse projection to world space. Aggregating across all pixels and views yields the scene-level representation $\hat{\mathcal{G}}$, which is rendered by a differentiable renderer $\mathcal{R}:(\hat{\mathcal{G}},\mathcal{T}^n)\mapsto \bar{\mathcal{I}}^n$, giving the rendered multi-view set $\bar{\mathcal{I}}=\{\bar{\mathcal{I}}^n\}_{n=1}^{N}$.

\paragraph{FM rendering and depth objectives.}
To align the rendering supervision with the velocity-based parameterization introduced in Section~\ref{sec:prelim}, we formulate the rendering loss in the same FM style. For each noisy view $n\in\Omega_{\mathrm n}$, we define the rendered and ground-truth velocities as $\bar{v}^n = (\bar{\mathcal{I}}^n-\tilde{\mathcal{I}}_t^n)/(1-t)$ and $v^n = (\mathcal{I}^n-\tilde{\mathcal{I}}_t^n)/(1-t)$, where $v^n=\mathcal{I}^n-\epsilon^n$ follows directly from Eq.~\eqref{eq:mixed_input}. For each clean view $n\in\Omega_{\mathrm c}$, the rendered image $\bar{\mathcal{I}}^n$ is supervised against the observation $\mathcal{I}^n$ by a direct MSE term, since clean views serve as noise-free reconstruction anchors. We then define
\begin{equation}
\mathcal{L}_{\mathrm{render}}
=
\underbrace{\frac{1}{|\Omega_{\mathrm c}|}\sum_{n\in\Omega_{\mathrm c}}
\|\bar{\mathcal{I}}^n-\mathcal{I}^n\|_2^2}_{\mathcal{L}_{\mathrm{recon}}}
+
\underbrace{\frac{\mathbf{1}[\,|\Omega_{\mathrm n}|>0\,]}{|\Omega_{\mathrm n}|}\sum_{n\in\Omega_{\mathrm n}}
\|\bar{v}^n-v^n\|_2^2}_{\mathcal{L}_{\mathrm{FM}}}
+
\lambda_{\mathrm{lpips}}\,\mathcal{L}_{\mathrm{lpips}},
\label{eq:lrender}
\end{equation}
where the FM term is equivalently written in image space as $
\frac{1}{|\Omega_{\mathrm n}|}\sum_{n\in\Omega_{\mathrm n}}
\frac{\|\bar{\mathcal{I}}^n-\mathcal{I}^n\|_2^2}{(1-t)^2}$,
and the perceptual term is defined as $\mathcal{L}_{\mathrm{lpips}}=\mathbf{1}[t>t_{\mathrm{th}}]\,\frac{1}{N}\sum_{n=1}^{N}\mathrm{LPIPS}(\bar{\mathcal{I}}^n,\mathcal{I}^n)$. The indicator $\mathbf{1}[\,|\Omega_{\mathrm n}|>0\,]$ disables the FM term in the pure reconstruction case ($\Omega_{\mathrm n}=\varnothing$), while $|\Omega_{\mathrm c}|\ge 1$ is guaranteed by Eq.~\eqref{eq:clean_noisy_partition}. We activate the LPIPS term only when $t>t_{\mathrm{th}}$, since perceptual supervision is unreliable when the noisy input is too close to pure noise. We further sample 4–8 novel views per iteration and supervise their renderings against ground-truth images via MSE and lpips loss, regularizing the 3DGS to render well from all viewpoints.

We further supervise the predicted depth maps against pseudo-depth labels $\mathcal{D}^{\star}=\{\mathcal{D}^{\star n}\}_{n=1}^{N}$ (obtained from DA3~\citep{lin2025depth}) by
\begin{equation}
\mathcal{L}_{\mathrm{depth}}
=
\frac{1}{N}\sum_{n=1}^{N}
\rho\!\left(
\log \hat{\mathcal{D}}^{\,n}
-
\log \mathcal{D}^{\star n}
\right),
\label{eq:ldepth}
\end{equation}
where $\rho(\cdot)$ is a Huber loss applied element-wise and then averaged over pixels.

%======================================================================
\subsection{Geometry Perception Loss}
\label{sec:geo_loss}
%======================================================================

\paragraph{Motivation.}
Although $\mathcal{L}_{\mathrm{render}}$ enforces view consistency, it remains photometric in nature and therefore does not fully constrain the underlying 3D structure. In particular, depth may drift along a viewing ray as long as the projected appearance is preserved, translucent floaters may average to plausible colors, and view-dependent texture can partially compensate for geometric misalignment. These ambiguities are especially problematic for noisy views, which do not have direct clean image supervision at the input stage.

To complement pixel-level supervision with a more structural signal, we introduce a \emph{geometry perception loss} defined in the feature space of a frozen 3D foundation model $\Psi$ (\textit{e.g.}, VGGT~\citep{wang2025vggt} or $\pi^3$~\citep{wang2025pi}).

\paragraph{Geometry perception loss.}
Given the rendered view set $\bar{\mathcal{I}}$ and the ground-truth view set $\mathcal{I}$ under the same cameras $\mathcal{T}$, we extract multi-view geometric features with $\Psi : (\mathcal{I},\mathcal{T}) \mapsto \mathcal{H}\in\mathbb{R}^{N\times C_{\Psi}\times H' \times W'}$. Applying $\Psi$ to the rendered and reference images gives
\begin{equation}
\bar{\mathcal{H}}=\Psi(\bar{\mathcal{I}},\mathcal{T})=\{\bar{\mathcal{H}}^n\}_{n=1}^{N},
\qquad
\mathcal{H}^{\star}=\Psi(\mathcal{I},\mathcal{T})=\{\mathcal{H}^{\star n}\}_{n=1}^{N},
\label{eq:geo_features}
\end{equation}
where $\bar{\mathcal{H}}^n$ and $\mathcal{H}^{\star n}$ denote the rendered and reference feature maps of the $n$-th view, respectively. We minimize the average per-location cosine distance between the two feature fields only when the timestep is sufficiently large:
\begin{equation}
\mathcal{L}_{\mathrm{geo}}
=
\mathbf{1}[t>t_{\mathrm{th}}]\,
\frac{1}{N H' W'}\sum_{n=1}^{N}\sum_{p}
\left[
1-
\frac{
\left\langle \bar{h}^{\,n}_{p},\,h^{\star n}_{p}\right\rangle
}{
\|\bar{h}^{\,n}_{p}\|_2\,\|h^{\star n}_{p}\|_2
}
\right],
\label{eq:geo_loss}
\end{equation}
where $\bar{h}^{\,n}_{p}$ and $h^{\star n}_{p}$ are the feature vectors at location $p$ of $\bar{\mathcal{H}}^n$ and $\mathcal{H}^{\star n}$, respectively. Because $\Psi$ jointly processes all views together with their cameras, its features encode cross-view 3D structure rather than only per-image appearance. Consequently, two renderings that are individually photo-consistent but mutually inconsistent in 3D can still produce different feature maps and incur a large $\mathcal{L}_{\mathrm{geo}}$. During training, we freeze $\Psi$, stop gradients on the reference branch $\mathcal{H}^{\star}$, and back-propagate only through the rendered branch $\bar{\mathcal{H}}$, so that $\Psi$ acts purely as a structural critic. We activate this term only when $t>t_{\mathrm{th}}$, since geometric feature matching is unstable when the noisy input is too close to pure noise.

\paragraph{Overall objective.}
The full training objective is $\mathcal{L} = \mathcal{L}_{\mathrm{render}} + \lambda_{\mathrm{depth}}\,\mathcal{L}_{\mathrm{depth}} + \lambda_{\mathrm{geo}}\,\mathcal{L}_{\mathrm{geo}}$.

\section{Experiments}

\begin{table}[t]
\centering
\caption{\textbf{Quantitative comparison of novel-view synthesis} on RealEstate10K~\citep{zhou2018stereo} and DL3DV-10K~\citep{ling2024dl3dv} under 4-view and 8-view input settings. Best results are in \textbf{bold}; second best are \underline{underlined}.}
\label{tab:nvs_comparison}
\setlength{\tabcolsep}{4pt}
\resizebox{\textwidth}{!}{%
\begin{tabular}{l ccc ccc ccc ccc}
\toprule
& \multicolumn{6}{c}{\textbf{RealEstate10K}} & \multicolumn{6}{c}{\textbf{DL3DV-10K}} \\
\cmidrule(lr){2-7} \cmidrule(lr){8-13}
& \multicolumn{3}{c}{4-views} & \multicolumn{3}{c}{8-views} & \multicolumn{3}{c}{4-views} & \multicolumn{3}{c}{8-views} \\
\cmidrule(lr){2-4} \cmidrule(lr){5-7} \cmidrule(lr){8-10} \cmidrule(lr){11-13}
Method
& PSNR$\uparrow$ & SSIM$\uparrow$ & LPIPS$\downarrow$
& PSNR$\uparrow$ & SSIM$\uparrow$ & LPIPS$\downarrow$
& PSNR$\uparrow$ & SSIM$\uparrow$ & LPIPS$\downarrow$
& PSNR$\uparrow$ & SSIM$\uparrow$ & LPIPS$\downarrow$ \\
\midrule
MVSplat~\citep{chen2024mvsplat}     
& 22.58 & 0.762 & 0.264 
& 21.64 & 0.719 & 0.301 
& 17.11 & 0.501 & 0.410 
& 15.75 & 0.432 & 0.491 \\
DepthSplat~\citep{xu2025depthsplat} 
& 25.16 & 0.832 & 0.194 
& 27.77 & 0.872 & 0.154 
& 20.38 & \textbf{0.719} & 0.320 
& 19.26 & \textbf{0.692} & 0.360 \\
AnySplat~\citep{jiang2025anysplat}  
& 20.07 & 0.731 & 0.286 
& 20.52 & 0.752 & 0.262 
& 20.11 & 0.671 & 0.318 
& 20.02 & 0.664 & 0.327 \\
YoNoSplat~\citep{ye2026yonosplat}   
& \underline{25.86} & \underline{0.841} & \underline{0.143} 
& \underline{28.35} & \underline{0.889} & \underline{0.107} 
& \underline{22.89} & 0.710 & \underline{0.228} 
& \underline{21.92} & 0.678 & \underline{0.262} \\
\rowcolor{cyan!8}
\textbf{PixWorld (Ours)}           
& \textbf{26.21} & \textbf{0.844} & \textbf{0.138} 
& \textbf{28.58} & \textbf{0.892} & \textbf{0.101} 
& \textbf{23.18} & \underline{0.714} & \textbf{0.226} 
& \textbf{22.46} & \underline{0.681} & \textbf{0.257} \\
\bottomrule
\end{tabular}%
}
\end{table}

\begin{table*}[t]
\centering
\caption{\textbf{Quantitative comparison on single-image 3D scene generation, averaged.} Results on RealEstate10K~\citep{zhou2018stereo} and DL3DV-10K~\citep{ling2024dl3dv} under the 1-view setting, averaged over First Frame and Bidirectional configurations. Best in \textbf{bold}; second best \underline{underlined}.}
\vspace{-2mm}
\label{tab:single_view_gen_avg}
\setlength{\tabcolsep}{4pt}
\renewcommand{\arraystretch}{1.05}
\resizebox{\textwidth}{!}{%
\begin{tabular}{l ccc cccc ccc}
\toprule
& \multicolumn{3}{c}{\textit{Novel View Synthesis}} & \multicolumn{4}{c}{\textit{Generation Quality}} & \multicolumn{3}{c}{\textit{Camera Control}} \\
\cmidrule(lr){2-4} \cmidrule(lr){5-8} \cmidrule(lr){9-11}
Method
& PSNR$\uparrow$ & SSIM$\uparrow$ & LPIPS$\downarrow$
& I2V Subj.$\uparrow$ & I2V BG$\uparrow$ & I.Q.$\uparrow$ & Aes.Q.$\uparrow$
& AUC@30$\uparrow$ & AUC@15$\uparrow$ & AUC@5$\uparrow$ \\
\midrule
\rowcolor{gray!12}
\multicolumn{11}{l}{\textbf{RealEstate10K}} \\
LVSM~\citep{jinlvsm}                & \underline{17.82} & 0.603 & \underline{0.336} & 0.971 & 0.970 & 0.593 & 0.506 & 0.710 & 0.592 & 0.372 \\
GF~\citep{wu2025geometry}           & 15.63 & 0.553 & 0.454 & 0.931 & 0.941 & 0.504 & 0.475 & 0.596 & 0.478 & 0.290 \\
Gen3C~\citep{ren2025gen3c}          & 17.26 & 0.624 & 0.391 & 0.951 & 0.956 & 0.561 & 0.524 & 0.648 & 0.514 & 0.334 \\
FlashWorld~\citep{li2025flashworld} & 16.51 & 0.626 & 0.403 & 0.958 & 0.960 & \underline{0.615} & \underline{0.550} & \underline{0.843} & \underline{0.758} & \underline{0.546} \\
Gen3R~\citep{huang2026gen3r}        & 17.59 & \underline{0.631} & 0.382 & \underline{0.974} & \underline{0.971} & 0.552 & 0.536 & 0.633 & 0.433 & 0.147 \\
\rowcolor{cyan!8}
\textbf{PixWorld (Ours)}           & \textbf{18.88} & \textbf{0.702} & \textbf{0.325} & \textbf{0.979} & \textbf{0.978} & \textbf{0.623} & \textbf{0.556} & \textbf{0.869} & \textbf{0.798} & \textbf{0.614} \\
\midrule
\rowcolor{gray!12}
\multicolumn{11}{l}{\textbf{DL3DV-10K}} \\
LVSM~\citep{jinlvsm}                & 14.91 & 0.433 & 0.530 & 0.931 & 0.933 & 0.494 & 0.466 & 0.552 & 0.372 & 0.134 \\
GF~\citep{wu2025geometry}           & 12.69 & 0.356 & 0.591 & 0.898 & 0.910 & 0.474 & 0.435 & 0.491 & 0.338 & 0.113 \\
Gen3C~\citep{ren2025gen3c}          & 15.58 & \underline{0.514} & 0.479 & 0.927 & 0.933 & 0.532 & 0.496 & 0.552 & 0.377 & 0.128 \\
FlashWorld~\citep{li2025flashworld} & 15.42 & 0.473 & \underline{0.461} & 0.942 & \underline{0.950} & \underline{0.619} & \underline{0.558} & \underline{0.769} & \underline{0.674} & \underline{0.420} \\
Gen3R~\citep{huang2026gen3r}        & \underline{15.75} & 0.503 & 0.495 & \underline{0.944} & 0.942 & 0.547 & 0.530 & 0.593 & 0.398 & 0.117 \\
\rowcolor{cyan!8}
\textbf{PixWorld (Ours)}           & \textbf{16.50} & \textbf{0.527} & \textbf{0.449} & \textbf{0.952} & \textbf{0.956} & \textbf{0.631} & \textbf{0.567} & \textbf{0.793} & \textbf{0.706} & \textbf{0.485} \\
\bottomrule
\end{tabular}}
\vspace{-2mm}
\end{table*}
\begin{table*}[t]
\centering
\caption{\textbf{Quantitative comparison on two-view 3D scene generation, averaged.} Results on RealEstate10K~\citep{zhou2018stereo} and DL3DV-10K~\citep{ling2024dl3dv} under the 2-view setting, averaged over Interpolation and Extrapolation configurations. Best in \textbf{bold}; second best \underline{underlined}.}
\vspace{-2mm}
\label{tab:two_view_gen_avg}
\setlength{\tabcolsep}{4pt}
\renewcommand{\arraystretch}{1.05}
\resizebox{\textwidth}{!}{%
\begin{tabular}{l ccc cccc ccc}
\toprule
& \multicolumn{3}{c}{\textit{Novel View Synthesis}} 
& \multicolumn{4}{c}{\textit{Generation Quality}} 
& \multicolumn{3}{c}{\textit{Camera Control}} \\
\cmidrule(lr){2-4} \cmidrule(lr){5-8} \cmidrule(lr){9-11}
Method
& PSNR$\uparrow$ & SSIM$\uparrow$ & LPIPS$\downarrow$
& I2V Subj.$\uparrow$ & I2V BG$\uparrow$ & I.Q.$\uparrow$ & Aes.Q.$\uparrow$
& AUC@30$\uparrow$ & AUC@15$\uparrow$ & AUC@5$\uparrow$ \\
\midrule
\rowcolor{gray!12}
\multicolumn{11}{l}{\textbf{RealEstate10K}} \\
LVSM~\citep{jinlvsm}                
& \textbf{23.61} & \textbf{0.819} & \underline{0.215} 
& 0.970 & 0.964 & 0.607 & 0.516 
& 0.861 & 0.788 & 0.611 \\
GF~\citep{wu2025geometry}           
& 18.27 & 0.647 & 0.353 
& 0.925 & 0.939 & 0.507 & 0.464 
& 0.630 & 0.473 & 0.223 \\
Gen3C~\citep{ren2025gen3c}          
& 20.12 & 0.714 & 0.300 
& 0.948 & 0.947 & 0.567 & 0.518 
& 0.698 & 0.538 & 0.255 \\
FlashWorld~\citep{li2025flashworld} 
& 21.48 & 0.770 & 0.257 
& 0.964 & 0.962 & \underline{0.619} & \underline{0.547} 
& \underline{0.877} & \underline{0.811} & \underline{0.637} \\
Gen3R~\citep{huang2026gen3r}        
& 21.33 & 0.724 & 0.283 
& \underline{0.970} & \underline{0.972} & 0.550 & 0.540 
& 0.728 & 0.576 & 0.258 \\
\rowcolor{cyan!8}
\textbf{PixWorld (Ours)}           
& \underline{23.54} & \underline{0.815} & \textbf{0.210} 
& \textbf{0.974} & \textbf{0.974} & \textbf{0.628} & \textbf{0.561} 
& \textbf{0.880} & \textbf{0.817} & \textbf{0.649} \\
\midrule
\rowcolor{gray!12}
\multicolumn{11}{l}{\textbf{DL3DV-10K}} \\
LVSM~\citep{jinlvsm}                
& \underline{19.18} & \underline{0.589} & \underline{0.343} 
& 0.915 & 0.917 & 0.533 & 0.502 
& 0.740 & 0.609 & 0.374 \\
GF~\citep{wu2025geometry}           
& 15.38 & 0.459 & 0.470 
& 0.897 & 0.912 & 0.479 & 0.445 
& 0.563 & 0.379 & 0.147 \\
Gen3C~\citep{ren2025gen3c}          
& 17.62 & 0.542 & 0.412 
& 0.927 & 0.934 & 0.536 & 0.502 
& 0.627 & 0.433 & 0.176 \\
FlashWorld~\citep{li2025flashworld} 
& 18.27 & 0.562 & 0.359 
& 0.938 & \underline{0.948} & \underline{0.600} & \underline{0.558} 
& \underline{0.802} & \underline{0.714} & \underline{0.514} \\
Gen3R~\citep{huang2026gen3r}        
& 18.05 & 0.558 & 0.392 
& \underline{0.942} & 0.944 & 0.535 & 0.530 
& 0.726 & 0.560 & 0.245 \\
\rowcolor{cyan!8}
\textbf{PixWorld (Ours)}           
& \textbf{19.37} & \textbf{0.594} & \textbf{0.340} 
& \textbf{0.950} & \textbf{0.956} & \textbf{0.607} & \textbf{0.565} 
& \textbf{0.821} & \textbf{0.734} & \textbf{0.534} \\
\bottomrule
\end{tabular}}
\vspace{-2mm}
\end{table*}

\begin{table*}[t]
\centering
\caption{\textbf{Quantitative comparison on the WorldScore benchmark~\citep{duan2025worldscore}.} We report all seven official metrics together with their average. \textbf{Bold} and \underline{underline} indicate the best and the second-best results, respectively.}
\vspace{-3mm}
\label{tab:worldscore}
\small
\renewcommand{\arraystretch}{1.0}
\setlength{\tabcolsep}{4pt}
\resizebox{\textwidth}{!}{%
\begin{tabular}{lccccccc|c}
\toprule
\textbf{Method}
& \makecell{Camera\\ Control}
& \makecell{Object\\ Control}
& \makecell{Content\\ Alignment}
& \makecell{3D\\ Consistency}
& \makecell{Photometric\\ Consistency}
& \makecell{Style\\ Consistency}
& \makecell{Subjective\\ Quality}
& \textbf{Average} \\
\midrule
Wan-2.1~\citep{wan2025wan}
& 23.53 & 40.32 & 45.44 & 78.74 & 78.36 & \underline{77.18} & \underline{59.38} & 57.56 \\
WonderJourney~\citep{yu2024wonderjourney}
& 84.60 & 37.10 & 35.54 & 80.60 & 79.03 & 62.82 & \textbf{66.56} & 63.75 \\
LucidDreamer~\citep{chung2023luciddreamer}
& \underline{88.93} & 41.18 & \textbf{75.00} & \underline{90.37} & \underline{90.20} & 48.10 & 58.99 & 70.40 \\
FlashWorld~\citep{li2025flashworld}
& 84.43 & \textbf{50.28} & \underline{56.54} & 85.87 & 86.72 & \textbf{79.36} & 52.75 & \underline{70.85} \\
\rowcolor{cyan!8}
\textbf{PixWorld (ours)}
& \textbf{91.08} & \underline{46.25} & 55.27 & \textbf{91.39} & \textbf{93.84} & 67.11 & 52.36 & \textbf{71.04} \\
\bottomrule
\end{tabular}%
}
\end{table*}

\subsection{Training Details}
PixWorld has $\sim\!1.04$B parameters and is trained from scratch on the Re10K~\citep{zhou2018stereo} + DL3DV-10K~\citep{ling2024dl3dv} mixture ($\sim\!67$K scenes in total), augmented with 10M single images from the BLIP-3o~\citep{chen2025blip3} corpus that share the diffusion backbone as a 2D appearance prior. For each multi-view scene we sample $N\!\in\!\{4,\ldots,8\}$ posed views and randomly partition them into $\Omega_{\mathrm c}$ and $\Omega_{\mathrm n}$, biased toward small $|\Omega_{\mathrm c}|$ so that capacity is spent on conditioned generation while the all-clean case ($\Omega_{\mathrm n}\!=\!\varnothing$) grounds the geometry head. We use $\Psi{=}\text{$\pi^3$}$~\citep{wang2025pi} as the frozen 3D critic for $\mathcal{L}_{\mathrm{geo}}$, and set $\lambda_{\mathrm{depth}}{=}1.0$ and $\lambda_{\mathrm{lpips}}{=}\lambda_{\mathrm{geo}}{=}0.1$, with the perceptual and geometric terms gated by $t{>}t_{\mathrm{th}}{=}0.3$. The model is optimized with AdamW~\citep{loshchilov2017decoupled} for $\sim\!200$K steps at a training resolution of $336\!\times\!448$ on $32$ NVIDIA A800-SXM4-80G GPUs. Full architectural specifications, batching, and optimizer schedules are deferred to the Appendix~\ref{sec:appendix_settings}.

\subsection{Evaluation Protocols}
\label{sec:eval_protocol}
\begin{figure}[t]
    \centering
    \vspace{-2mm}
    \includegraphics[width=\linewidth]{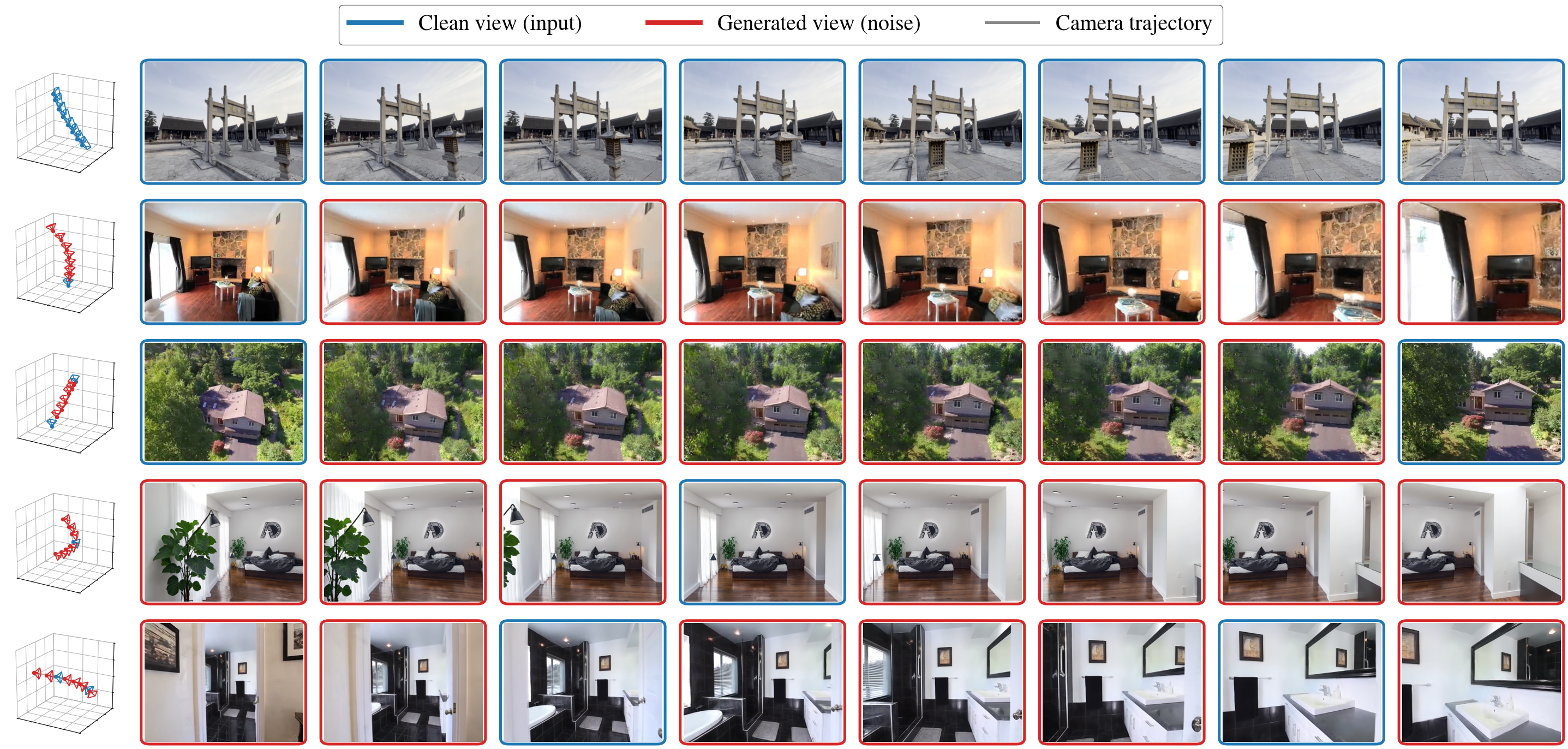}
    \caption{\textbf{Visualization of PixWorld under different settings.} PixWorld flexibly handles both 3D reconstruction and generation: when all input views are clean, it performs reconstruction; when clean and noisy views are arbitrarily mixed, it performs generation. We visualize the camera trajectory, where blue and red frustums denote clean input views and generated views, respectively.}
    \label{fig:pose_traj_vis}
    \vspace{-4mm}
\end{figure}

We evaluate PixWorld on four protocols covering 3D reconstruction and 3D scene generation (see Fig.~\ref{fig:pose_traj_vis}). For RealEstate10K~\citep{zhou2018stereo} and DL3DV-10K~\citep{ling2024dl3dv}, we randomly sample 200 test scenes per dataset, restricted to clips with a large camera pose range so the protocols stress wide-baseline reasoning; for WorldScore~\citep{duan2025worldscore} we follow the official static split (2000 scenes). For all scene-generation protocols, every baseline is conditioned on camera poses, and all baselines except LVSM~\citep{jinlvsm} additionally receive a text condition. For reconstruction (Tab.~\ref{tab:nvs_comparison}), the model is given 4 or 8 posed views and renders held-out targets under ground-truth poses, evaluated by PSNR, SSIM~\citep{wang2004image}, and LPIPS~\citep{zhang2018unreasonable}. For 1-view generation (Tab.~\ref{tab:single_view_gen_avg}), the 200 scenes per dataset are split into 100 First-Frame scenes (forward trajectory generated from the first frame) and 100 Bidirectional scenes (a randomly chosen middle frame conditions generation toward both ends); 2-view generation (Tab.~\ref{tab:two_view_gen_avg}) similarly combines 100 Interpolation scenes (endpoint anchors, intermediate views generated) and 100 Extrapolation scenes (anchors at one end, views generated beyond their span). For both generation settings we report three groups of metrics, each targeting one capability of a 3D world model: Novel View Synthesis (NVS) fidelity against ground-truth target views (PSNR, SSIM, LPIPS); Generation Quality from VBench~\citep{huang2024vbench} (I2V Subject, I2V Background, Image Quality, Aesthetic Quality), assessing perceptual realism without paired references; and Camera Control precision via $\pi^3$~\citep{wang2025pi}-estimated AUC@$\{30^\circ, 15^\circ, 5^\circ\}$ ~\citep{wang2025taming} between the poses recovered from generated frames and the conditioning trajectory. On WorldScore (Tab.~\ref{tab:worldscore}) we report all seven official metrics and their average: Camera Control and Object Control measure trajectory control fidelity, Content Alignment measures faithfulness to the text prompt, 3D and Photometric Consistency assess geometric and appearance stability across views, Style Consistency captures visual style coherence, and Subjective Quality reflects overall perceptual quality, with baseline numbers taken from the WorldScore release.

\subsection{3D Scene Reconstruction}
\label{sec:reconstruction}

Tab.~\ref{tab:nvs_comparison} reports novel-view synthesis results on RealEstate10K and DL3DV-10K under 4-view and 8-view inputs. Note that Gen3R~\citep{huang2026gen3r}, although a unified generation-and-reconstruction model, only supports point-cloud reconstruction and is therefore not directly comparable on NVS, so it is omitted here. We further note that YoNoSplat supports both a pose-free and a pose-conditioned (\textit{with-pose}) mode; for a stronger comparison we adopt its with-pose results, which leverage ground-truth camera poses as additional input. Even against this stronger baseline, PixWorld attains the best PSNR and LPIPS across all settings, as well as the best SSIM on RealEstate10K, trailing only DepthSplat on SSIM for DL3DV-10K. Concretely, PixWorld consistently improves PSNR over YoNoSplat on both RealEstate10K and DL3DV-10K (4/8 views), and lowers LPIPS in every case (\textit{e.g.},\ $0.138$ vs.\ $0.143$ at 4-view RealEstate10K). These results show that our pixel-space formulation and geometry-aware supervision yield stronger cross-view consistency and more accurate 3D reconstruction.

\subsection{3D Scene Generation}
\label{subsec:scene_gen}
\begin{table}[t]
\centering
\caption{\textbf{Ablation study on geometry perception loss.}
We report results on RealEstate10K~\citep{zhou2018stereo} under the \emph{1-view} setting.}
\label{tab:ablation_geometry_perception}
\setlength{\tabcolsep}{4pt}
\renewcommand{\arraystretch}{1.05}
\resizebox{\columnwidth}{!}{%
\begin{tabular}{l ccc cccc ccc}
\toprule
Variant
& PSNR$\uparrow$ & SSIM$\uparrow$ & LPIPS$\downarrow$
& I2V Subj.$\uparrow$ & I2V BG$\uparrow$ & I.Q.$\uparrow$ & Aes.Q.$\uparrow$
& AUC@30$\uparrow$ & AUC@15$\uparrow$ & AUC@5$\uparrow$ \\
\midrule
Full model 
& 19.12 & 0.717 & 0.310
& 0.972 & 0.975 & 0.619 & 0.561 
& 0.886 & 0.813 & 0.642 \\

w/o Geometry Perception 
& 17.99 & 0.612 & 0.332
& 0.973 & 0.974 & 0.613 & 0.541 
& 0.847 & 0.763 & 0.562  \\
\bottomrule
\end{tabular}}
\vspace{-3mm}
\end{table}
\begin{figure}[t]
    \centering
    \includegraphics[width=\linewidth]{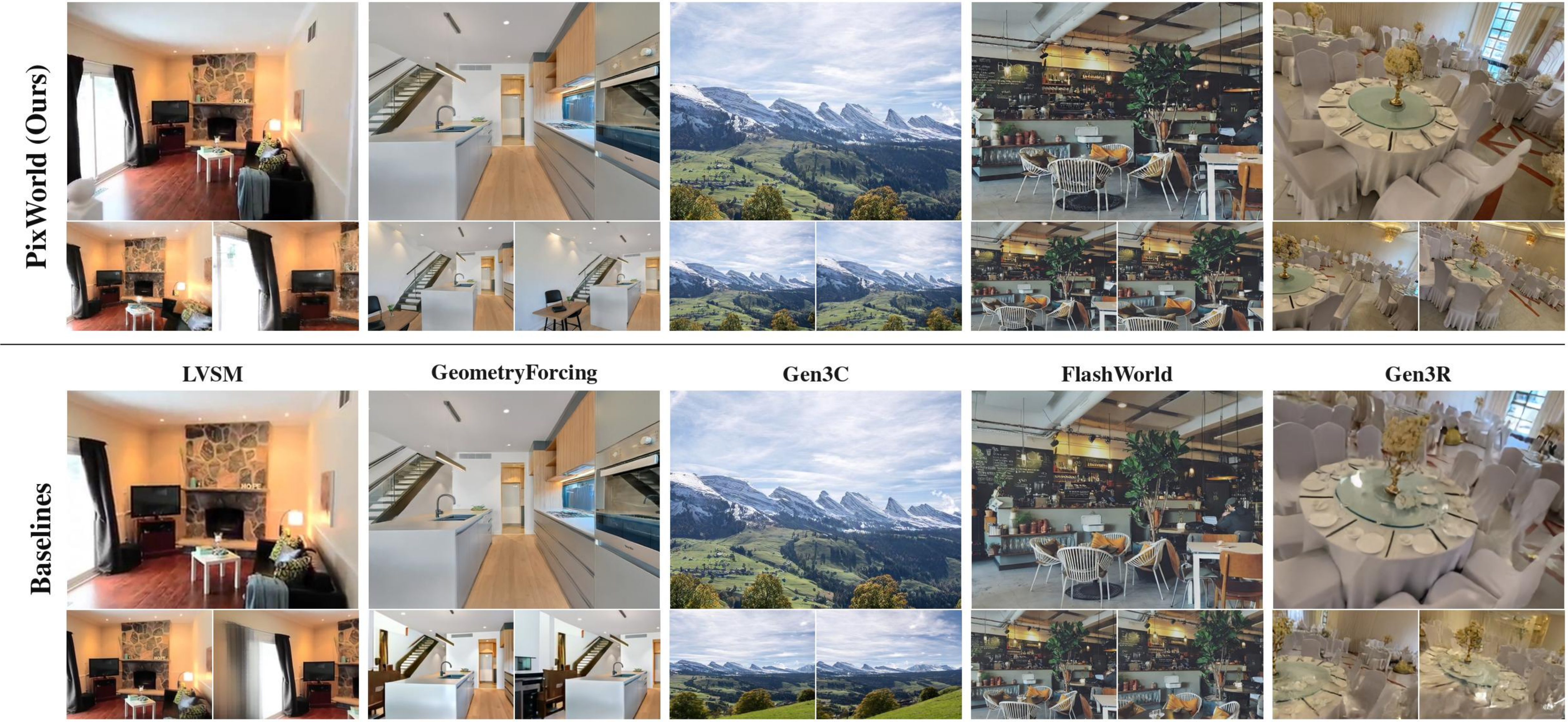}
    \caption{\textbf{Visualization of comparisons with baselines.} The large view on top denotes the input view, while the two smaller views below show novel views generated by each method.}
    \vspace{-4mm}
    \label{fig:vis_compare_baselines}
\end{figure}

We benchmark PixWorld against five representative baselines, LVSM~\citep{jinlvsm}, GF~\citep{wu2025geometry}, Gen3C~\citep{ren2025gen3c}, FlashWorld~\citep{li2025flashworld} and Gen3R~\citep{huang2026gen3r}, on RealEstate10K~\citep{zhou2018stereo} and DL3DV-10K~\citep{ling2024dl3dv} under single and two-image conditioning. We report novel-view synthesis quality (PSNR, SSIM, LPIPS), VBench-style generation quality, and pose accuracy via $\pi^3$-estimated AUC at multiple thresholds. In the single-image setting (Tab.~\ref{tab:single_view_gen_avg}), averaged across First-Frame and Bidirectional trajectories, PixWorld tops every metric on both datasets, lifting PSNR by $+1.06$\,dB on RealEstate10K (18.88 vs.\ 17.82) and $+0.75$\,dB on DL3DV-10K (16.50 vs.\ 15.75); the gain is sharpest at strict pose thresholds, with AUC@5 rising from 0.546 to 0.614 and from 0.420 to 0.485, indicating that diffusing in pixel space and supervising through differentiable rendering yields geometrically faithful trajectories. Under two-image conditioning (Tab.~\ref{tab:two_view_gen_avg}), PixWorld again leads on perceptual, generation, and pose metrics, attaining the best LPIPS (0.210/0.340) and AUC@5 (0.649/0.534); LVSM is competitive only on raw PSNR/SSIM (gap $<0.07$\,dB on RealEstate10K), reflecting its deterministic regression objective. In WorldScore~\citep{duan2025worldscore} (Tab.~\ref{tab:worldscore}), PixWorld achieves the best overall average (71.04) and ranks first in camera controllability (91.08), 3D consistency (91.39) and photometric consistency (93.84); see Fig.~\ref{fig:vis_compare_baselines}.

\subsection{Ablation Study}
\label{subsec:ablation}

We ablate the geometry perception loss on RealEstate10K~\citep{zhou2018stereo} under the 1-view setting (Tab.~\ref{tab:ablation_geometry_perception}). For a controlled comparison, we sample a 10K-sequence subset and train both variants for 30K steps under identical settings, toggling only the geometry perception loss. Removing it consistently degrades all three metric groups: PSNR drops by $1.13$\,dB ($19.12 \!\to\! 17.99$), SSIM by $0.105$ ($0.717 \!\to\! 0.612$) and AUC@5 by $0.080$ ($0.642 \!\to\! 0.562$, $\sim$$12.5\%$ relative), while VBench-style scores barely shift. This pattern matches our motivation: 2D photometric and perceptual losses keep individual frames visually plausible but leave the underlying 3D geometry unsupervised, so cross-view consistency and pose fidelity suffer (see Fig.~\ref{fig:vis_ablation_study} in Appendix~\ref{sec:appendix_vis}). By aligning rendered and ground-truth views in the geometry-aware feature space of a pretrained 3D foundation model, our loss supplies the 3D structural signal that 2D objectives cannot, validating it as a key component of PixWorld.
\section{Conclusion}
We present PixWorld, an end-to-end pixel-space diffusion framework that unifies 3D scene generation and reconstruction in a single model by partitioning multi-view inputs into clean and noisy subsets and producing a pixel-aligned 3D Gaussian representation in one forward pass. Eliminating the intermediate VAE/RAE stage avoids the information loss and extra training cost of latent autoencoders, and lets the diffusion objective directly supervise the 3D representation through differentiable rendering rather than an intermediate latent target; to further enforce 3D structural consistency beyond 2D photometric and perceptual losses, a geometry perception loss aligns rendered and ground-truth views in the geometry-aware feature space of a pretrained 3D foundation model. These results suggest that pixel-space diffusion, by removing latent indirection and naturally unifying generation with reconstruction, marks a promising paradigm toward scalable and unified 3D scene modeling.

\bibliography{main}
\bibliographystyle{iclr2026_conference}

\clearpage
\appendix
\section{Overview of the Appendix}
This appendix supplements the main paper along five axes. Appendix~\ref{sec:appendix_settings} details the architecture of the two-stream MMDiT denoiser with a per-component parameter budget, together with the data, batching, and optimization recipes used to train PixWorld from scratch. Appendix~\ref{sec:appendix_per_config} provides disaggregated quantitative results on 1-view and 2-view generation, separating the configurations averaged in the main paper to reveal how each method behaves on the harder versus easier side of each setting. Appendix~\ref{sec:appendix_vis} presents additional qualitative visualizations covering reconstruction and generation under diverse view selections, with both RGB renderings and predicted depth maps. Appendix~\ref{sec:appendix_speed} reports an inference-speed comparison against representative baselines. Finally, the responsible-considerations (Appendix~\ref{appendix:responsible}) section discusses limitations, broader impacts, and LLM usage.

\section{Implementation Details}
\label{sec:appendix_settings}

\paragraph{Architecture.}
PixWorld's denoiser $f_\theta$ is a 24-layer DiT~\citep{peebles2023scalable} with hidden width $d{=}1024$, $16$ attention heads (head dim $64$), SwiGLU~\citep{shazeer2020glu} feed-forward layers, RMSNorm on $Q,K$, and adaLN-Zero modulation conditioned on the diffusion timestep. Following the SD3-style MMDiT~\citep{esser2024scaling} design, each block hosts two parallel streams that share topology but maintain independent pre-LayerNorm, QKV / output projections, MLP, and adaLN-Zero weights: a clean stream processes the conditioning views in $\Omega_{\mathrm c}$, and a noise stream processes the noisy views in $\Omega_{\mathrm n}$. Inside the attention operator, the per-stream $Q$, $K$, $V$ tensors are concatenated along the token axis (with shared $q,k$-RMSNorm) and a single full attention is computed jointly over $[\,\Omega_{\mathrm c};\,\Omega_{\mathrm n}\,]$, so conditioning and noisy tokens cross-talk in every layer; the joint output is then split back and routed through stream-specific output projections. The two streams share a single cross-attention to text tokens and a single timestep embedder, the latter evaluated at $t{=}1$ for the clean stream and at the sampled $t$ for the noise stream, allowing one model to serve both pose-only and text-conditioned generation. Camera parameters are injected via PRoPE~\citep{li2025cameras}. Inputs use a $16{\times}16$ patchify with learnable positional embeddings, and the final layer emits per-pixel depth and 3D-Gaussian attributes through stream-specific multi-task heads, with Gaussian centers obtained by unprojecting each pixel using its predicted depth (Sec.~\ref{sec:pixworld}). The total trainable parameter count is $1.044$\,B; Table~\ref{tab:arch} reports the per-component budget.

\begin{table}[t]
\centering
\caption{\textbf{Architecture of the two-stream DiT denoiser $f_\theta$.} Each block follows an MMDiT-style design: the clean and noise streams maintain independent pre-LayerNorm, QKV / output projections, SwiGLU MLP, and adaLN-Zero weights, while a single full attention is computed jointly over the concatenated $[\,\Omega_{\mathrm c};\,\Omega_{\mathrm n}\,]$ tokens with shared $q,k$-RMSNorm. The cross-attention to text and the timestep embedder are also shared across streams, and output heads are duplicated per stream so that both clean and noisy tokens are decoded into depth and 3D-Gaussian attributes at every patch.}
\label{tab:arch}
\small
\setlength{\tabcolsep}{5pt}
\renewcommand{\arraystretch}{1.05}
\begin{tabular}{@{}l l r@{}}
\toprule
Module & Configuration & \#Params \\
\midrule
\multicolumn{3}{@{}l}{(a) Tokenization \& conditioning} \\
Clean-view patch embed     & $16{\times}16$ patchify, $3 {\rightarrow} 1024$            & 0.79\,M \\
Noisy-view patch embed     & $16{\times}16$ patchify, $3 {\rightarrow} 1024$            & 0.79\,M \\
Positional embedding       & learnable, $588 {\times} 1024$                              & 0.60\,M \\
Timestep embedder          & sinusoidal $256$ + MLP $1024 {\rightarrow} 1024$            & 1.31\,M \\
Text projection            & MLP $4096 {\rightarrow} 1024 {\rightarrow} 1024$            & 5.24\,M \\
\midrule
\multicolumn{3}{@{}l}{(b) Two-stream MMDiT block ($\times\,24$;\, $d{=}1024$, $h{=}16$, $d_h{=}64$)} \\
QKV projection (clean)         & pre-LN + $1024 {\rightarrow} 3{\times}1024$                  & 3.15\,M \\
Output projection (clean)      & $1024 {\rightarrow} 1024$                                    & 1.05\,M \\
SwiGLU MLP (clean)             & $1024 {\rightarrow} 2{\times}2730 {\rightarrow} 1024$        & 8.39\,M \\
adaLN-Zero (clean)             & $t$-cond.\ $\gamma,\beta,\alpha$ ($\times 6$)                & 6.30\,M \\
QKV projection (noise)         & pre-LN + $1024 {\rightarrow} 3{\times}1024$                  & 3.15\,M \\
Output projection (noise)      & $1024 {\rightarrow} 1024$                                    & 1.05\,M \\
SwiGLU MLP (noise)             & $1024 {\rightarrow} 2{\times}2730 {\rightarrow} 1024$        & 8.39\,M \\
adaLN-Zero (noise)             & $t$-cond.\ $\gamma,\beta,\alpha$ ($\times 6$)                & 6.30\,M \\
Joint full attention           & SDPA over $[\Omega_{\mathrm c};\Omega_{\mathrm n}]$ with shared $q,k$-RMSNorm & $\sim 0$ \\
Cross-attention to text        & shared by both streams                                       & 4.20\,M \\
\multicolumn{2}{r}{per-block subtotal}                                                & 41.98\,M \\
\multicolumn{2}{r}{trunk total ($\times 24$ blocks)}                                  & 1007.6\,M \\
\midrule
\multicolumn{3}{@{}l}{(c) Multi-task output heads (duplicated per stream)} \\
Depth head       & adaLN + linear $1024 {\rightarrow} 16{\times}16{\times}1$  & $2{\times}2.36$\,M \\
3D-Gaussian head & adaLN + linear $1024 {\rightarrow} 16{\times}16{\times}35$ & $2{\times}11.28$\,M \\
\midrule
Total trainable parameters & & 1.044\,B \\
\bottomrule
\end{tabular}
\end{table}

\paragraph{Data and batching.}
The RealEstate10K~\citep{zhou2018stereo} and DL3DV-10K~\citep{ling2024dl3dv} datasets together provide $\sim\!67$K posed multi-view scenes. From each scene we sample $N\!\in\!\{4,\ldots,8\}$ views and partition them into $\Omega_{\mathrm c}/\Omega_{\mathrm n}$, biasing the sampler toward small $|\Omega_{\mathrm c}|$ so capacity is spent on conditioned generation. We additionally mix in a single-image branch of 10M images drawn from the BLIP-3o~\citep{chen2025blip3} corpus, which shares the diffusion backbone and strengthens the 2D appearance prior. On each GPU, every optimization step alternates a 32-image single-view batch with a multi-view batch of up to 32 images (\textit{e.g.}, $4$--$8$ views $\times$ $4$--$8$ scenes).

\paragraph{Optimization.}
We optimize $\mathcal{L} = \mathcal{L}_{\mathrm{render}} + \lambda_{\mathrm{depth}}\,\mathcal{L}_{\mathrm{depth}} + \lambda_{\mathrm{geo}}\,\mathcal{L}_{\mathrm{geo}}$ with $\lambda_{\mathrm{depth}}{=}1.0$ and $\lambda_{\mathrm{lpips}}{=}\lambda_{\mathrm{geo}}{=}0.1$, gating $\mathcal{L}_{\mathrm{lpips}}$ and $\mathcal{L}_{\mathrm{geo}}$ at $t{>}t_{\mathrm{th}}{=}0.3$ since perceptual and geometric supervision is unreliable when the noisy input is close to pure noise. The frozen geometry critic $\Psi$ is instantiated as $\pi^3$~\citep{wang2025pi}, with gradients stopped on the reference branch. Training is run from scratch (no image- or video-model pretraining) using AdamW~\citep{loshchilov2017decoupled} with a linearly decayed learning rate from $1\!\times\!10^{-4}$ to $1\!\times\!10^{-5}$, EMA decay $0.9995$, gradient clipping at $1.0$, and classifier-free text dropping at rate $0.2$. Training runs for $\sim\!200$K steps on $32$ NVIDIA A800-SXM4-80G GPUs.
\section{Detailed Results on 1-View and 2-View Generation}
\label{sec:appendix_per_config}

\begin{table*}[t]
\centering
\caption{\textbf{Quantitative comparison on single-image 3D scene generation, per configuration.} We evaluate on RealEstate10K~\citep{zhou2018stereo} and DL3DV-10K~\citep{ling2024dl3dv} under 1-view First Frame and 1-view Bidirectional. Best in \textbf{bold}; second best \underline{underlined}.}
\label{tab:single_view_gen}
\setlength{\tabcolsep}{4pt}
\renewcommand{\arraystretch}{1.05}
\resizebox{\textwidth}{!}{%
\begin{tabular}{l ccc cccc ccc}
\toprule
& \multicolumn{3}{c}{\textit{Novel View Synthesis}} & \multicolumn{4}{c}{\textit{Generation Quality}} & \multicolumn{3}{c}{\textit{Camera Control}} \\
\cmidrule(lr){2-4} \cmidrule(lr){5-8} \cmidrule(lr){9-11}
Method
& PSNR$\uparrow$ & SSIM$\uparrow$ & LPIPS$\downarrow$
& I2V Subj.$\uparrow$ & I2V BG$\uparrow$ & I.Q.$\uparrow$ & Aes.Q.$\uparrow$
& AUC@30$\uparrow$ & AUC@15$\uparrow$ & AUC@5$\uparrow$\\
\midrule
\rowcolor{gray!12}
\multicolumn{11}{l}{\textbf{RealEstate10K}~~--~~1-view First Frame} \\
LVSM~\citep{jinlvsm}                & \underline{17.95} & 0.604 & \underline{0.335} & 0.968 & \underline{0.971} & 0.584 & 0.505 & 0.695 & 0.576 & 0.359 \\
GF~\citep{wu2025geometry}           & 15.92 & 0.575 & 0.436 & 0.942 & 0.947 & 0.514 & 0.487 & 0.609 & 0.488 & 0.301 \\
Gen3C~\citep{ren2025gen3c}          & 17.12 & 0.622 & 0.394 & 0.959 & 0.961 & 0.569 & 0.530 & 0.664 & 0.521 & 0.338 \\
FlashWorld~\citep{li2025flashworld} & 16.26 & 0.613 & 0.417 & 0.951 & 0.954 & \textbf{0.617} & \underline{0.544} & \underline{0.849} & \underline{0.766} & \underline{0.553} \\
Gen3R~\citep{huang2026gen3r}        & 17.43 & \underline{0.628} & 0.383 & \underline{0.973} & 0.970 & 0.547 & 0.531 & 0.653 & 0.443 & 0.145 \\
\rowcolor{cyan!8}
\textbf{PixWorld (Ours)}           & \textbf{18.92} & \textbf{0.683} & \textbf{0.324} & \textbf{0.977} & \textbf{0.979} & \underline{0.608} & \textbf{0.549} & \textbf{0.872} & \textbf{0.804} & \textbf{0.621} \\
\midrule
\rowcolor{gray!12}
\multicolumn{11}{l}{\textbf{RealEstate10K}~~--~~1-view Bidirectional} \\
LVSM~\citep{jinlvsm}                & 17.70 & 0.601 & \underline{0.337} & 0.974 & 0.969 & 0.603 & 0.506 & 0.726 & 0.608 & 0.385 \\
GF~\citep{wu2025geometry}           & 15.34 & 0.532 & 0.471 & 0.921 & 0.935 & 0.495 & 0.462 & 0.583 & 0.467 & 0.279 \\
Gen3C~\citep{ren2025gen3c}          & 17.39 & 0.626 & 0.388 & 0.943 & 0.952 & 0.553 & 0.517 & 0.632 & 0.507 & 0.329 \\
FlashWorld~\citep{li2025flashworld} & 16.75 & \underline{0.639} & 0.390 & 0.966 & 0.967 & \underline{0.614} & \underline{0.555} & \underline{0.837} & \underline{0.751} & \underline{0.539} \\
Gen3R~\citep{huang2026gen3r}        & \underline{17.74} & 0.633 & 0.381 & \underline{0.975} & \underline{0.972} & 0.557 & 0.541 & 0.612 & 0.424 & 0.150 \\
\rowcolor{cyan!8}
\textbf{PixWorld (Ours)}           & \textbf{18.83} & \textbf{0.721} & \textbf{0.325} & \textbf{0.981} & \textbf{0.978} & \textbf{0.639} & \textbf{0.563} & \textbf{0.865} & \textbf{0.793} & \textbf{0.607} \\
\midrule
\rowcolor{gray!12}
\multicolumn{11}{l}{\textbf{DL3DV-10K}~~--~~1-view First Frame} \\
LVSM~\citep{jinlvsm}                & 14.85 & 0.434 & 0.533 & 0.927 & 0.932 & 0.491 & 0.463 & 0.537 & 0.354 & 0.127 \\
GF~\citep{wu2025geometry}           & 12.50 & 0.352 & 0.598 & 0.894 & 0.907 & 0.468 & 0.431 & 0.482 & 0.329 & 0.109 \\
Gen3C~\citep{ren2025gen3c}          & 15.72 & \underline{0.516} & 0.482 & 0.925 & 0.928 & 0.526 & 0.491 & 0.556 & 0.371 & 0.121 \\
FlashWorld~\citep{li2025flashworld} & 15.31 & 0.468 & \underline{0.463} & 0.939 & \underline{0.950} & \underline{0.616} & \underline{0.558} & \underline{0.765} & \underline{0.667} & \underline{0.412} \\
Gen3R~\citep{huang2026gen3r}        & \underline{15.57} & 0.499 & 0.504 & \underline{0.941} & 0.940 & 0.543 & 0.529 & 0.614 & 0.408 & 0.102 \\
\rowcolor{cyan!8}
\textbf{PixWorld (Ours)}           & \textbf{16.37} & \textbf{0.521} & \textbf{0.455} & \textbf{0.947} & \textbf{0.953} & \textbf{0.624} & \textbf{0.564} & \textbf{0.789} & \textbf{0.702} & \textbf{0.477} \\
\midrule
\rowcolor{gray!12}
\multicolumn{11}{l}{\textbf{DL3DV-10K}~~--~~1-view Bidirectional} \\
LVSM~\citep{jinlvsm}                & 14.96 & 0.432 & 0.526 & 0.934 & 0.933 & 0.497 & 0.468 & 0.568 & 0.391 & 0.141 \\
GF~\citep{wu2025geometry}           & 12.88 & 0.361 & 0.584 & 0.902 & 0.913 & 0.479 & 0.439 & 0.501 & 0.346 & 0.118 \\
Gen3C~\citep{ren2025gen3c}          & 15.44 & \underline{0.512} & 0.476 & 0.929 & 0.937 & 0.537 & 0.501 & 0.548 & 0.382 & 0.134 \\
FlashWorld~\citep{li2025flashworld} & 15.53 & 0.478 & \underline{0.458} & 0.944 & \underline{0.950} & \underline{0.621} & \underline{0.559} & \underline{0.773} & \underline{0.681} & \underline{0.428} \\
Gen3R~\citep{huang2026gen3r}        & \underline{15.94} & 0.507 & 0.487 & \underline{0.948} & 0.944 & 0.551 & 0.532 & 0.571 & 0.389 & 0.133 \\
\rowcolor{cyan!8}
\textbf{PixWorld (Ours)}           & \textbf{16.63} & \textbf{0.533} & \textbf{0.443} & \textbf{0.956} & \textbf{0.958} & \textbf{0.637} & \textbf{0.571} & \textbf{0.797} & \textbf{0.709} & \textbf{0.493} \\
\bottomrule
\end{tabular}}
\end{table*}
\begin{table*}[t]
\centering
\caption{\textbf{Quantitative comparison on two-view 3D scene generation, per configuration.} We evaluate on RealEstate10K~\citep{zhou2018stereo} and DL3DV-10K~\citep{ling2024dl3dv} under 2-view Interpolation and 2-view Extrapolation. Best in \textbf{bold}; second best \underline{underlined}.}
\label{tab:two_view_gen}
\setlength{\tabcolsep}{4pt}
\renewcommand{\arraystretch}{1.05}
\resizebox{\textwidth}{!}{%
\begin{tabular}{l ccc cccc ccc}
\toprule
& \multicolumn{3}{c}{\textit{Novel View Synthesis}} & \multicolumn{4}{c}{\textit{Generation Quality}} & \multicolumn{3}{c}{\textit{Camera Control}} \\
\cmidrule(lr){2-4} \cmidrule(lr){5-8} \cmidrule(lr){9-11}
Method
& PSNR$\uparrow$ & SSIM$\uparrow$ & LPIPS$\downarrow$
& I2V Subj.$\uparrow$ & I2V BG$\uparrow$ & I.Q.$\uparrow$ & Aes.Q.$\uparrow$
& AUC@30$\uparrow$ & AUC@15$\uparrow$ & AUC@5$\uparrow$\\
\midrule
\rowcolor{gray!12}
\multicolumn{11}{l}{\textbf{RealEstate10K}~~--~~2-view Interpolation} \\
LVSM~\citep{jinlvsm}                & \textbf{24.26} & \textbf{0.832} & \textbf{0.200} & \underline{0.969} & 0.963 & 0.613 & 0.518 & 0.871 & 0.806 & \underline{0.638} \\
GF~\citep{wu2025geometry}           & 18.08 & 0.621 & 0.373 & 0.936 & 0.951 & 0.525 & 0.476 & 0.642 & 0.487 & 0.236 \\
Gen3C~\citep{ren2025gen3c}          & 20.17 & 0.726 & 0.298 & 0.967 & 0.949 & 0.581 & 0.527 & 0.714 & 0.553 & 0.261 \\
FlashWorld~\citep{li2025flashworld} & 21.66 & 0.776 & 0.252 & 0.958 & 0.959 & \underline{0.619} & \underline{0.546} & \underline{0.875} & \underline{0.810} & 0.634 \\
Gen3R~\citep{huang2026gen3r}        & 22.42 & 0.752 & 0.260 & 0.964 & \underline{0.973} & 0.544 & 0.539 & 0.779 & 0.631 & 0.295 \\
\rowcolor{cyan!8}
\textbf{PixWorld (Ours)}           & \underline{23.44} & \underline{0.810} & \underline{0.214} & \textbf{0.971} & \textbf{0.975} & \textbf{0.625} & \textbf{0.559} & \textbf{0.877} & \textbf{0.814} & \textbf{0.648} \\
\midrule
\rowcolor{gray!12}
\multicolumn{11}{l}{\textbf{RealEstate10K}~~--~~2-view Extrapolation} \\
LVSM~\citep{jinlvsm}                & \underline{22.95} & \underline{0.806} & \underline{0.230} & 0.972 & 0.965 & 0.601 & 0.514 & 0.851 & 0.771 & 0.583 \\
GF~\citep{wu2025geometry}           & 18.46 & 0.672 & 0.332 & 0.914 & 0.927 & 0.489 & 0.451 & 0.618 & 0.458 & 0.211 \\
Gen3C~\citep{ren2025gen3c}          & 20.08 & 0.703 & 0.302 & 0.928 & 0.945 & 0.552 & 0.509 & 0.682 & 0.524 & 0.249 \\
FlashWorld~\citep{li2025flashworld} & 21.30 & 0.765 & 0.261 & 0.970 & 0.965 & \underline{0.618} & \underline{0.548} & \underline{0.878} & \underline{0.813} & \underline{0.640} \\
Gen3R~\citep{huang2026gen3r}        & 20.24 & 0.695 & 0.307 & \underline{0.976} & \underline{0.970} & 0.556 & 0.542 & 0.677 & 0.521 & 0.220 \\
\rowcolor{cyan!8}
\textbf{PixWorld (Ours)}           & \textbf{23.63} & \textbf{0.819} & \textbf{0.206} & \textbf{0.978} & \textbf{0.974} & \textbf{0.631} & \textbf{0.564} & \textbf{0.883} & \textbf{0.819} & \textbf{0.651} \\
\midrule
\rowcolor{gray!12}
\multicolumn{11}{l}{\textbf{DL3DV-10K}~~--~~2-view Interpolation} \\
LVSM~\citep{jinlvsm}                & \textbf{19.37} & \textbf{0.594} & \textbf{0.328} & 0.918 & 0.917 & 0.538 & 0.511 & 0.758 & 0.630 & 0.403 \\
GF~\citep{wu2025geometry}           & 15.16 & 0.452 & 0.480 & 0.903 & 0.918 & 0.487 & 0.452 & 0.573 & 0.395 & 0.156 \\
Gen3C~\citep{ren2025gen3c}          & 17.81 & 0.546 & 0.406 & 0.931 & 0.938 & 0.541 & 0.508 & 0.641 & 0.446 & 0.183 \\
FlashWorld~\citep{li2025flashworld} & 18.16 & 0.559 & 0.363 & 0.937 & \underline{0.951} & \underline{0.602} & \underline{0.564} & \underline{0.812} & \underline{0.724} & \underline{0.528} \\
Gen3R~\citep{huang2026gen3r}        & 18.21 & 0.562 & 0.391 & \underline{0.940} & 0.946 & 0.531 & 0.536 & 0.745 & 0.582 & 0.260 \\
\rowcolor{cyan!8}
\textbf{PixWorld (Ours)}           & \underline{19.12} & \underline{0.581} & \underline{0.348} & \textbf{0.953} & \textbf{0.957} & \textbf{0.611} & \textbf{0.568} & \textbf{0.826} & \textbf{0.739} & \textbf{0.542} \\
\midrule
\rowcolor{gray!12}
\multicolumn{11}{l}{\textbf{DL3DV-10K}~~--~~2-view Extrapolation} \\
LVSM~\citep{jinlvsm}                & \underline{18.98} & \underline{0.584} & \underline{0.358} & 0.913 & 0.916 & 0.528 & 0.493 & 0.723 & 0.587 & 0.346 \\
GF~\citep{wu2025geometry}           & 15.61 & 0.466 & 0.459 & 0.891 & 0.906 & 0.471 & 0.438 & 0.553 & 0.362 & 0.137 \\
Gen3C~\citep{ren2025gen3c}          & 17.42 & 0.539 & 0.418 & 0.924 & 0.930 & 0.532 & 0.496 & 0.614 & 0.421 & 0.169 \\
FlashWorld~\citep{li2025flashworld} & 18.37 & 0.565 & \underline{0.356} & 0.938 & \underline{0.946} & \underline{0.598} & \underline{0.552} & \underline{0.793} & \underline{0.703} & \underline{0.501} \\
Gen3R~\citep{huang2026gen3r}        & 17.88 & 0.554 & 0.392 & \underline{0.943} & 0.941 & 0.539 & 0.523 & 0.706 & 0.538 & 0.231 \\
\rowcolor{cyan!8}
\textbf{PixWorld (Ours)}           & \textbf{19.61} & \textbf{0.607} & \textbf{0.331} & \textbf{0.947} & \textbf{0.954} & \textbf{0.604} & \textbf{0.561} & \textbf{0.817} & \textbf{0.728} & \textbf{0.526} \\
\bottomrule
\end{tabular}}
\end{table*}

\begin{figure}[t]
    \centering
    \includegraphics[width=\linewidth]{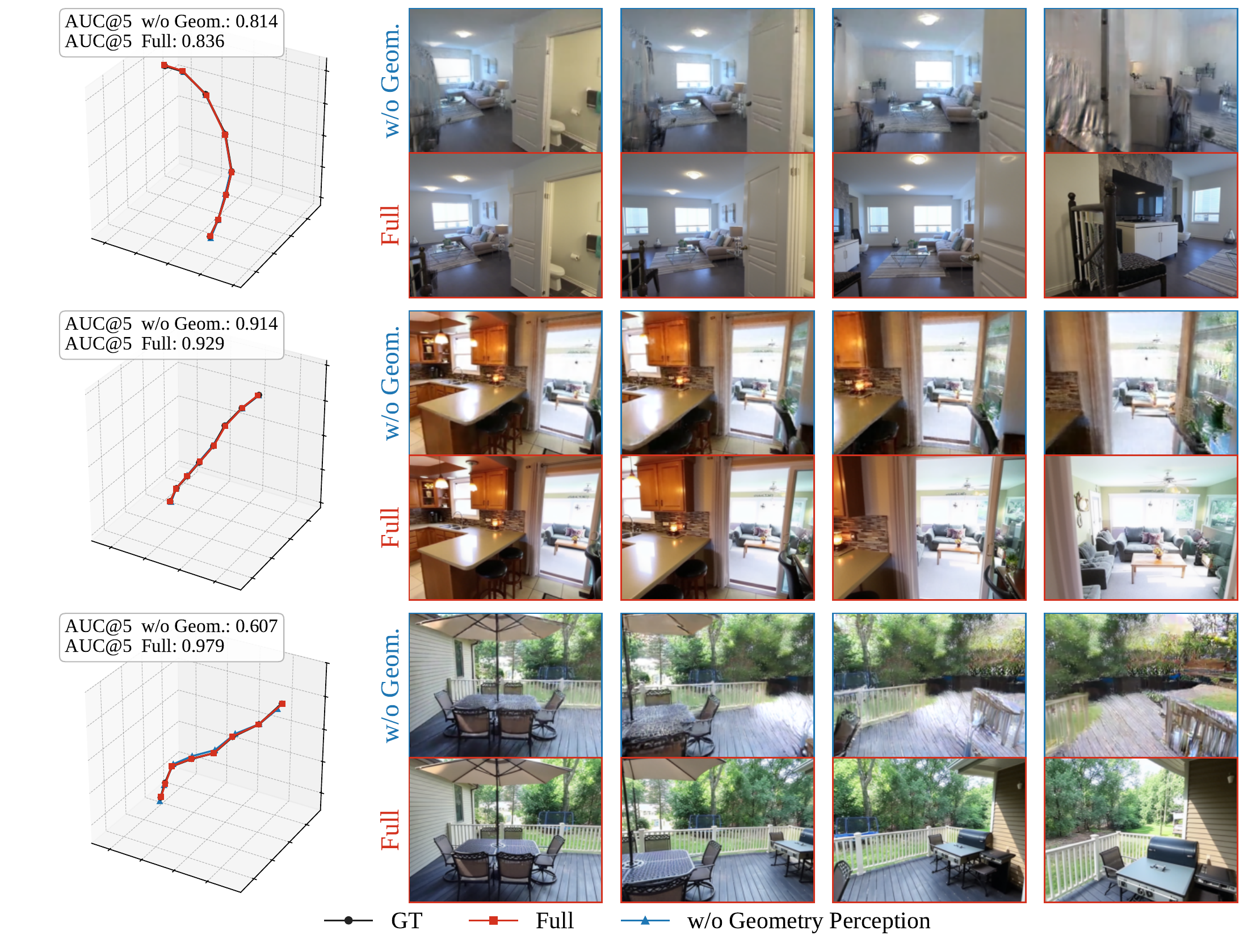}
    \caption{\textbf{Ablation study on the Geometry Perception loss in PixWorld.} Given a single input image, our model generates the subsequent 7 frames (8 frames in total); we visualize 4 representative frames here for clarity. Pose accuracy is quantitatively evaluated by comparing the estimated camera poses against the ground-truth poses. Compared to the variant without Geometry Perception (\textit{w/o Geom.}), the full model achieves more precise camera pose control and substantially mitigates the blurriness in later-view predictions, demonstrating that the global 3D perception loss is essential for maintaining both geometric consistency and visual fidelity over long generation horizons.}
    \label{fig:vis_ablation_study}
\end{figure}

The main paper (Tab.~\ref{tab:single_view_gen_avg} and Tab.~\ref{tab:two_view_gen_avg}) reports numbers averaged within each input setting to keep the comparison compact. For completeness, we provide the disaggregated per-configuration results here. Under the 1-view setting, the average is taken over two complementary configurations: First Frame, where the input view is the first frame of the clip and the model generates a purely forward trajectory along a long, fully extrapolative horizon; and Bidirectional, where a randomly chosen middle frame conditions generation toward both ends, producing two shorter and roughly symmetric horizons that more directly test local consistency around the input. Under the 2-view setting, the average is taken over Interpolation, where the two anchors bracket the target trajectory and the model fills in intermediate views, and Extrapolation, where both anchors lie at one end of the clip and the model must generate views beyond their span under stronger parallax. The disaggregated results in Tab.~\ref{tab:single_view_gen} and Tab.~\ref{tab:two_view_gen} reveal two consistent trends. First, every method degrades on the harder side of each setting (First Frame for 1-view, Extrapolation for 2-view), with the largest gaps appearing on LPIPS and on the Camera Control AUC@$\{30^\circ, 15^\circ, 5^\circ\}$ metrics. Second, the method ranking is largely preserved across configurations, confirming that the averaged numbers in the main paper faithfully summarize each method's behavior rather than being dominated by the easier sub-configuration. PixWorld remains the top performer on nearly every column across all four configurations, with the largest margins precisely on the harder sides, indicating that our design generalizes best in the extrapolation- and parallax-heavy regimes where competing methods suffer most.

\section{Additional Visualizations}
\label{sec:appendix_vis}
\begin{figure}[t]
    \centering
    \includegraphics[width=\linewidth]{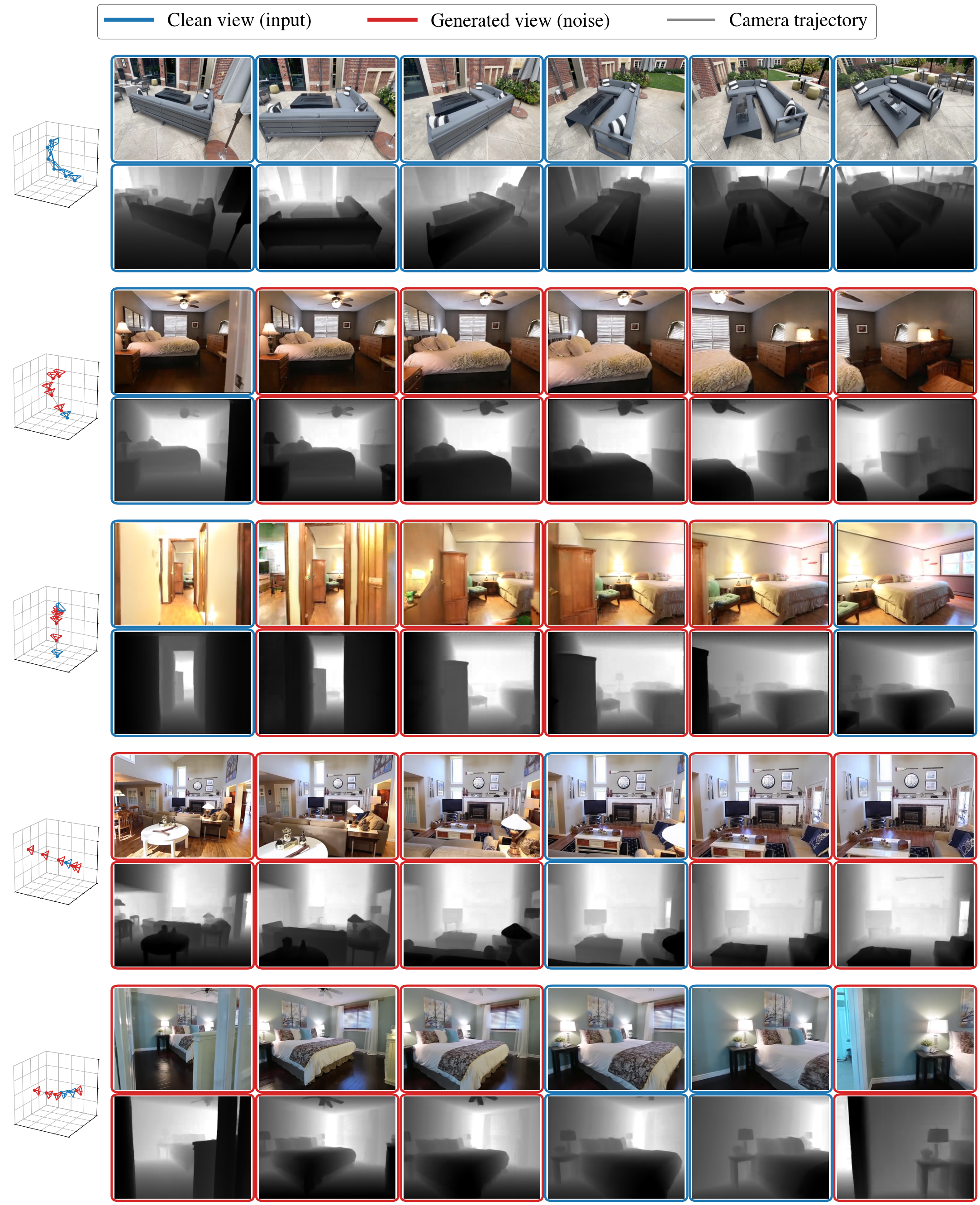}
    \caption{More visualizations of reconstruction and generation under varying view selections, including camera trajectories with input and generated views marked, and the corresponding depth maps predicted by PixWorld.}
    \label{fig:appendix_vis1}
\end{figure}

\begin{figure}[t]
    \centering
    \includegraphics[width=\linewidth]{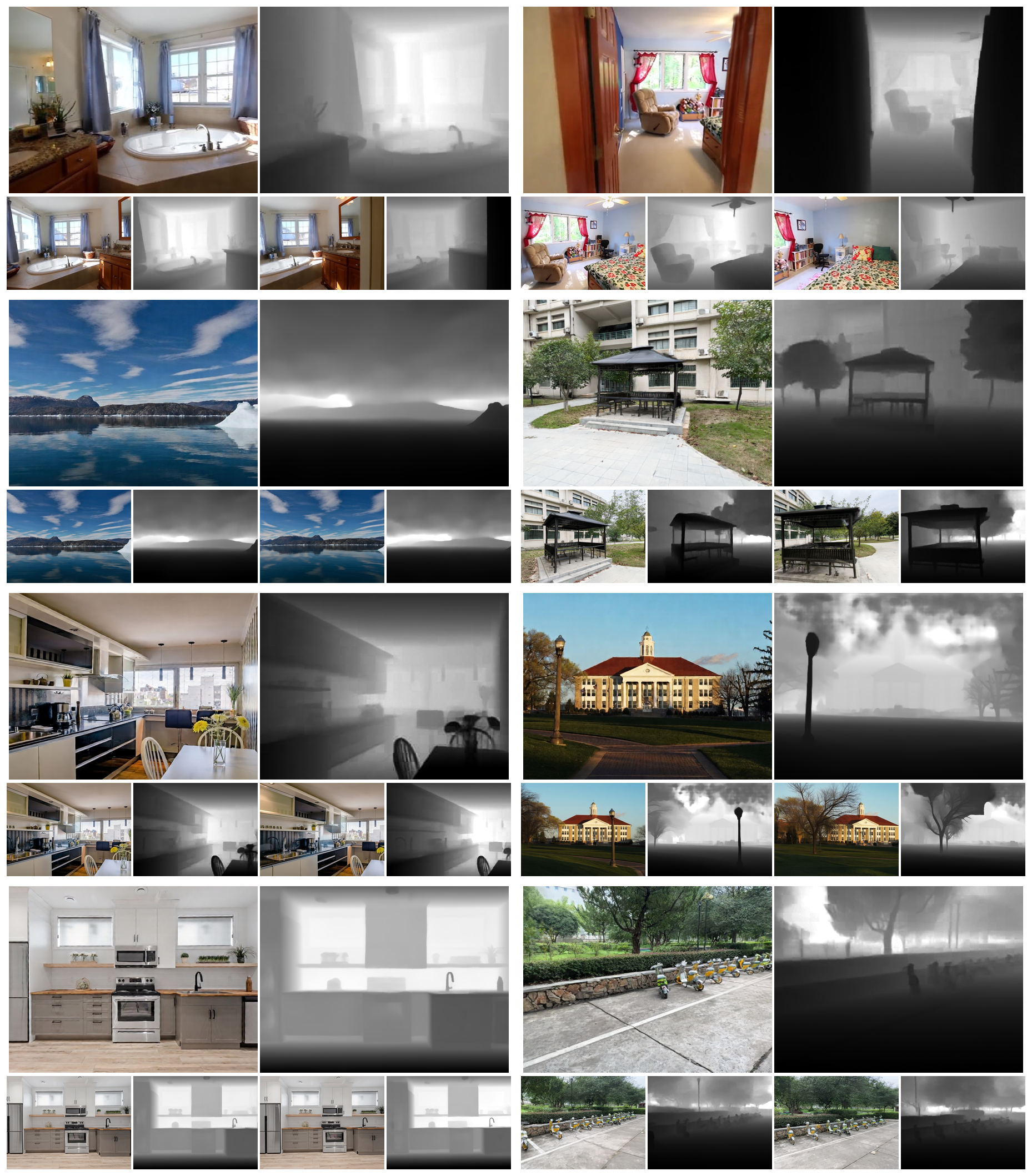}
    \caption{More visualizations of generated scenes. The first view is the input, and we show both RGB renderings and predicted depth maps.}
    \label{fig:appendix_vis2}
\end{figure}

We provide additional qualitative results to complement the quantitative comparisons in the main paper. Fig.~\ref{fig:appendix_vis1} shows reconstruction and generation across diverse view selections: for each scene, we visualize the camera trajectory with input views and generated views marked, alongside the RGB renderings and the depth maps predicted by PixWorld. These examples cover varied input counts and trajectory shapes, and demonstrate that PixWorld produces geometrically coherent scenes regardless of how the input views are arranged. Fig.~\ref{fig:appendix_vis2} further showcases generated scenes from a single input view (shown as the first frame of each sequence), with both RGB renderings and the corresponding depth maps. Across these examples, the predicted depth remains sharp and structurally consistent with the rendered appearance, indicating that the joint depth and 3D-Gaussian prediction in PixWorld captures scene geometry faithfully even under heavy extrapolation from a single conditioning image. Finally, Fig.~\ref{fig:vis_ablation_study} complements our ablation study by qualitatively comparing PixWorld with and without the Geometry Perception loss, where the full model yields sharper later-view renderings and more accurate pose control. Beyond these on-page visualizations, we further provide video comparisons of 3D scenes generated by different methods in the supplementary zip file, where the camera pose of each rendered video is additionally estimated to quantitatively assess camera control precision.

\begin{table}[t]
\centering
\caption{\textbf{Inference speed comparison} on a single NVIDIA A100-SXM4-80G GPU. We report the wall-clock time to generate one scene, the number of key frames per scene, and the number of function evaluations (NFE).}
\label{tab:inference_speed}
\setlength{\tabcolsep}{8pt}
\renewcommand{\arraystretch}{1.05}
\begin{tabular}{l c c c}
\toprule
Method & \#Key Frames per scene & NFE & Time per scene (s) $\downarrow$ \\
\midrule
Gen3C~\citep{ren2025gen3c}          & 121 & 70  & 791 \\
Gen3R~\citep{huang2026gen3r}        & 49  & 100 & 882 \\
FlashWorld~\citep{li2025flashworld} & 24  & 4   & \textbf{10} \\
\rowcolor{cyan!8}
\textbf{PixWorld (Ours)}           & 8   & 100 & \underline{15} \\
\bottomrule
\end{tabular}
\end{table}

\section{Inference Speed Comparison}
\label{sec:appendix_speed}
We benchmark inference speed on a single NVIDIA A100-SXM4-80G GPU in Tab.~\ref{tab:inference_speed}, reporting wall-clock time per scene, the number of key frames, and the number of function evaluations (NFE). PixWorld generates a scene in 15 seconds, reaching the same order of magnitude as the highly optimized FlashWorld (10\,s). We note that this comparison is not strictly apples-to-apples: PixWorld currently runs at a lower output resolution than the video-diffusion baselines, which reduces per-step compute. More fundamentally, PixWorld does not rely on video-model priors and thus does not need to materialize a dense frame sequence: a 3D representation is reconstructed from as few as 8 key frames, after which novel views are rendered efficiently via differentiable rasterization. By contrast, video-diffusion pipelines denoise a long frame sequence end-to-end, with Gen3C, Gen3R, and FlashWorld producing 121, 49, and 24 frames per scene, respectively. Note also that PixWorld and Gen3R use a comparable NFE (100), while FlashWorld leverages distillation to reduce its NFE to 4, which largely accounts for its strong runtime. We view distillation as complementary to our approach: the numbers above are for the base, undistilled PixWorld, and combining it with distillation~\citep{yin2024one,yin2024improved} and post-training quantization~\citep{li2023q,shang2023post} is a natural next step (see Appendix~\ref{apdx:limitations}).

\section{Responsible Considerations}
\label{appendix:responsible}

\subsection{Limitations.}
\label{apdx:limitations}

PixWorld takes a step toward unifying 3D scene reconstruction and generation in pixel space, but several directions remain open. Our experiments focus on widely used scene-level datasets such as RealEstate10K and DL3DV-10K; further evaluation on more diverse outdoor and object-centric scenes would better characterize the framework's generalization. Pixel-space diffusion with differentiable rendering is also trained under finite resolution and compute budgets, leaving room for improvements in fine-grained texture fidelity and scalability to higher-resolution multi-view settings. Finally, we plan to accelerate PixWorld's inference through distillation~\citep{yin2024one,yin2024improved} and quantization~\citep{li2023q,shang2023post}, further reducing the cost of high-quality 3D scene generation.

\subsection{Broader impacts.}
\label{apdx:broader_impact}
PixWorld may benefit applications such as efficient 3D reconstruction, 3D content creation, robotic perception, simulation, and VR/AR. At the same time, improved 3D scene reconstruction and generation may raise concerns about privacy-sensitive scene capture, misuse of synthetic or reconstructed 3D content, and unreliable deployment in safety-critical settings. We encourage responsible data usage, careful evaluation, and human oversight when applying such systems in real-world scenarios.
\subsection{LLM usage.}
\label{apdx:llm_usage}
This work does not use LLMs as an important, original, or non-standard component of the core method. Any use of LLM-based tools, if any, was limited to writing, editing, or formatting assistance and did not affect the methodology, experiments, or scientific conclusions.

\end{document}